\documentclass[sigconf,10pt]{acmart}

\usepackage{pifont}
\usepackage{balance}
\usepackage{enumitem}
\usepackage{graphicx}
\usepackage{subfigure}
\usepackage{multirow}

\usepackage{tablefootnote}
\usepackage{threeparttable}

\usepackage[ruled, vlined, linesnumbered, ruled]{algorithm2e}
\usepackage[noend]{algpseudocode}

\usepackage{color}
\usepackage{xspace}
\usepackage{fancyhdr}
\usepackage{multicol}
\usepackage{amsfonts}
\usepackage{graphicx}
\usepackage{tabularx}
\usepackage{threeparttable}
\usepackage{makecell}
\usepackage{multicol}
\usepackage{multirow}
\usepackage{xparse}
\usepackage{arydshln}

\usepackage{bbm} % mathbb 1

\usepackage{bm}

\usepackage{tikz}
\newcommand*{\circled}[1]{\lower.8ex\hbox{\tikz\draw (0pt, 0pt)%
    circle (.47em) node {\makebox[0.4em][c]{\small #1}};}}

\def\ie{\textit{i.e.}\xspace}

\def\eg{\textit{e.g.}\xspace}

\def\oursys{{HierFedLoRA}\xspace}

\setcopyright{none}
\begin{document}
 
% \title{\bluenote{Efficient Federated Fine-tuning Pre-trained Language Models on Heterogeneous Devices}}
% \title{Tacking Heterogeneity Issues in Parameter-Efficient Federated Fine-Tuning Large Language Model}
% \title{\rednote{Parameter-Efficient Fine-Tuning for Heterogeneous Data on Devices}}
% \title{{Resource-Efficient Federated Fine-Tuning Large Language Models for Heterogeneous Data}}
\title{Resource-Efficient Federated Fine-Tuning Large Language Models for Heterogeneous Data}

% \author{Submission ID \#434}
% Your Institutio
% \and
% {\rm Second Name}\\
% Second Institution
% copy the following lines to add more authors
% \and
% {\rm Name}\\
%Name Institution
% } % end author
\author{\Large
Jun Liu$^{1,2}$, Yunming Liao$^{1,2}$, Hongli Xu$^{1,2}$, Yang Xu$^{1,2}$\\
$^{1}$School of Computer Science and Technology, University of Science and Technology of China\\
$^{2}$Suzhou Institute for Advanced Research, University of Science and Technology of China\\
% $^{3}$Department of Computer Science and Engineering, University of California at Santa Cruz\\
% \{lj18090759057, ymliao98, xuhongli, xuyangcs, jcliu17\}@mail.ustc.edu.cn,~\{cqian12\}@ucsc.edu
}

\begin{abstract}
Fine-tuning large language models (LLMs) via federated learning, \ie, FedLLM, has been proposed to adapt LLMs for various downstream applications in a privacy-preserving way.
To reduce the fine-tuning costs on resource-constrained devices, FedLoRA is proposed to fine-tune only a small subset of model parameters by integrating low-rank adaptation (LoRA) into FedLLM.
% To alleviate the high resource cost of full fine-tuning, low-rank adaptation (LoRA), which only fine-tunes a small number of model parameters, has been proposed and adopted in FedLLM, \ie, FedLoRA.
However, apart from resource constraints, there is still another critical challenge, \ie, data heterogeneity, severely hindering the implementation of FedLoRA in practical applications.
% existing works mainly focus on LoRA initialization and rank optimization but fail to conquer two critical challenges, \ie, data heterogeneity and resource constraints, which severely hinder convergence rate and efficiency.
Herein, inspired by the previous group-based federated learning paradigm, we propose a hierarchical FedLoRA framework, termed \oursys, to address these challenges.
Specifically, \oursys partitions all devices into multiple near-IID groups and adjusts the intra-group aggregation frequency for each group to eliminate the negative effects of non-IID data.
Meanwhile, to reduce the computation and communication cost, \oursys dynamically assigns diverse and suitable fine-tuning depth (\ie, the number of continuous fine-tuning layers from the output) for each group.
\oursys explores jointly optimizing aggregation frequency and depth upon their coupled relationship to better enhance the performance of FedLoRA.
% Comprehensive experiments on the physical platform show that \oursys enhances model accuracy by 1.6–4.2\% and accelerates fine-tuning by at least 2.1$\times$ compared to baselines.
Extensive experiments are conducted on a physical platform with 80 commercial devices. 
The results show that \oursys improves the final model accuracy by 1.6\% to 4.2\%, speeding up the fine-tuning process by at least 2.1$\times$, compared to the strong baselines.

\end{abstract} 




\maketitle
\thispagestyle{empty} % 去除页眉
\pagestyle{empty}

\pagenumbering{arabic}
\fancyfoot[C]{\fontsize{10pt}{40pt}\selectfont\thepage}
\setcounter{page}{1}
\thispagestyle{fancy}

%-------------------------------------------------------------------------------
\section{Introduction}
% \textbf{Intro of FedNLP.}
% Large language models (LLMs) are changing the landscape of modern applications. 
% The development of large language models (LLMs)  reshaping software development paradigms
% Recently, the rapid development of large language models (LLMs), \eg, GPT-4 \cite{achiam2023gpt} and Llama 3 \cite{grattafiori2024llama}, have revolutionized the field of natural language processing, leading to significant breakthroughs in various applications.
% Large language models (LLMs) such as GPT \cite{achiam2023gpt} and Llama \cite{grattafiori2024llama} are revolutionizing the landscape of modern applications, leading to significant breakthroughs in various applications.
The rapid advancement of large language models (LLMs), such as GPT \cite{achiam2023gpt} and Llama \cite{grattafiori2024llama}, has propelled the development of artificial intelligence, transforming the landscape of modern applications.
% Large language models (LLMs) have catalyzed significant advancements in natural language processing, unveiling the immense potential for various applications.
% As foundation models, their remarkable success has naturally sparked interest in fine-tuning for particular application scenarios, such as sentiment analysis, question answering, data retrieval, and beyond.
As foundation models, pre-trained LLMs can be adapted to various downstream tasks through fine-tuning and have been widely applied in mobile applications, including but not limited to sentiment analysis \cite{zhang2023sentiment}, question answering \cite{jiang2020can}, and personal assistance \cite{xu2018deeptype}.
% Large language models (LLMs) are revolutionizing the application ecosystem.
% Generally, LLMs such as GPT4 \cite{achiam2023gpt} and Llama2 \cite{touvron2023llama} are first pre-trained on a large corpus to learn general features and patterns.
% Then, the pre-trained LLMs are fine-tuned to particular application scenarios, \eg, fine-tuning Llama2 for better code generation \cite{roziere2023code}.
% Generally, for most real-world applications, pre-trained LLMs are centrally fine-tuned using extensive datasets collected from a wide range of devices \cite{qu2021natural}.
Despite the promise of fine-tuning LLMs, there is a growing concern about collecting the necessary high-quality data for fine-tuning LLMs due to data privacy \cite{mcmahan2017communication, lin2021fednlp}, \eg, the European Union’s General Data Protection Regulation (GDPR) \footnote{https://gdpr-info.eu/}.

To this end, early efforts have focused on fine-tuning LLMs through federated learning, known as FedLLM \cite{lin2021fednlp, zhang2022federated, cai2022fedadapter, cho2023heterogeneous}, to fully utilize the massive data on devices.
In a traditional FedLLM framework, \eg, FedNLP \cite{lin2021fednlp}, devices fine-tune local LLMs on their data and periodically upload local LLMs to the parameter server (PS) for global aggregation, iterating until convergence or reaching the target accuracy.
% However, fine-tuning the entire LLM in FedLLM incurs significant memory, computing, and communication overheads on edge devices due to the inherent large size of LLMs.
However, due to the inherent large size of LLMs, fine-tuning the entire LLM in FedLLM incurs significant computation and communication overheads on the devices.
% 分为计算和通信来部分来讲，以常规的配置，说明时延，具体的值，可以收敛到计算才是瓶颈，通过增加通信可以使得计算高效
For instance, fine-tuning a Llama2-7B model \cite{touvron2023llama} on the device requires a computation cost exceeding 2.1 PetaFLOPs in a round, which may take several hours for a modern commercial device, \eg, Jetson AGX Xavier, to complete the local fine-tuning \cite{qin2023federated}.
Besides, transmitting the updated model parameters to the PS may also require several hours, depending on the network bandwidth \cite{mcmahan2017communication}.
% For instance, fine-tuning a Llama2-7B model \cite{touvron2023llama} on the device demands over 98.3GB of GPU memory \cite{smith2022using} and requires 12.3GB of network traffic for uploading the model updates to the server per round, which may separately take hours for local fine-tuning and transmission \cite{mcmahan2017communication}.
To tackle this issue, low-rank adaptation (LoRA) \cite{hu2021lora}, one of the most popular parameter-efficient fine-tuning methods \cite{houlsby2019parameter, hu2021lora, li2021prefix, zaken2021bitfit, zhao2022reduce, zhang2023fedpetuning}, has been proposed and widely adopted in FedLLM (called FedLoRA) \cite{zhang2022federated, cho2023heterogeneous, babakniya2023slora, bai2024federated}.  
FedLoRA reduces fine-tuning costs on the devices by updating/exchanging only a small subset of model parameters (typically less than 1\%) while keeping the pre-trained LLM unchanged. 
Compared to traditional FedLLM, FedLoRA greatly reduces communication costs by over 99\% (\ie, reducing transmission time to just a few seconds per round) while maintaining comparable performance \cite{hu2021lora, zhang2023lora, zhang2023adalora}.

\textbf{Challenges of FedLoRA.}
Although FedLoRA has demonstrated its advantages, it still suffers from two other critical challenges in practical applications: 
(1) \textit{Data heterogeneity.}
The devices always collect local data based on locations and user preferences, resulting in non-independent and identically distributed (non-IID) data across all devices \cite{lin2021fednlp, babakniya2023slora}.
The non-IID data will decelerate the convergence rate and even compromise the final performance of the fine-tuned LLM.
Besides, due to the limited number of tunable parameters, FedLoRA is susceptible to non-IID data \cite{zhang2023fedpetuning}.
(2) \textit{Resource constraints}. 
Many devices, such as personal computers and in-vehicle devices, typically have limited resources (\eg, computing power and bandwidth), which are orders of magnitude weaker than cloud servers \cite{zhu2023pockengine, dhar2021survey}. 
% Additionally, LoRA effectively reduces memory overhead but retains high computational costs, leading to slow fine-tuning on resource-constrained devices.
In addition, existing LLMs, \eg, Llama 2 \cite{touvron2023llama}, typically involve billions of parameters, requiring substantial computing power for updating the tunable parameters (even for LoRA \cite{hu2021lora,wu2024fedbiot}), while resource-constrained devices always lead to slow convergence rates.
% Additionally, existing LLMs, \eg, Llama2 \cite{touvron2023llama}, typically involve billions of parameters, requiring substantial computing power and memory for fine-tuning all layers (even for LoRA \cite{hu2021lora}), while resource-constrained devices always lead to very slow fine-tuning rates.
% (2) \textit{System heterogeneity.}
% The devices commonly possess varying and limited capacities. 
% The computing capacities (\eg, CPU frequency) and communication capacities (\eg, bandwidth) of devices could differ from each other by more than tenfold times \cite{chen2022decentralized, liu2023yoga}.
% Even the same type of devices with diverse configurations (\eg, smartphones) \cite{dhar2021survey}. 

\textbf{Status Quo and Limitations.}
Current research of FedLoRA primarily focuses on the intrinsic setup of LoRA, \eg, LoRA initialization \cite{babakniya2023slora, yan2024federa} or LoRA rank \cite{cho2023heterogeneous, bai2024federated, fang2024automated}, yet demonstrates critical limitations to address these challenges.
First, LoRA initialization is to initialize the LoRA parameters by decomposing the weight matrices of the pre-trained LLM.
However, it often results in degraded performance, especially in heterogeneous data scenarios where parameter drift contributes to degradation \cite{yan2024federa}.
% First, LoRA initialization is to initialize the LoRA parameters by decomposing the weight matrices of the pre-trained LLM, while it often results in degraded performance, especially in heterogeneous data scenarios where parameter drift contributes to degradation \cite{yan2024federa}.
% Moreover, such approaches are often model-dependent and may require fine-tuning the entire LLM, making them impractical for resource-constrained scenarios where devices cannot support full LLM fine-tuning \cite{babakniya2023slora}.
Second, assigning suitable LoRA rank for different devices improves the fine-tuning performance to some extent but fails to reduce the high computation cost of fine-tuning on the devices \cite{cho2023heterogeneous, bai2024federated}.
This is because LoRA fine-tuning still requires complete forward and backward propagation, resulting in computation costs comparable to full fine-tuning \cite{kaddour2023challenges}.
Consequently, they do not address the two above challenges.

\textbf{Overview of the Proposed Approach.}
% \rednote{
% Naturally, devices can be divided into multiple groups, each with a near-IID data distribution, to mitigate parameter drift between the group and global models. 
% Through hierarchical intra-group and inter-group aggregation, the non-IID issue can be effectively addressed.
% }
%Motivated by the insights from established federated learning paradigms \cite{lee2020accurate, he2022improving}, where 
Recalling the previous federated learning paradigms \cite{lee2020accurate, he2022improving}, devices can be divided into multiple groups, each with a near-IID data distribution, to mitigate parameter drift between the group and global models.
% Through hierarchical intra-group and inter-group aggregation, the non-IID issue can be effectively addressed.
% Given the parameter-efficient nature of LoRA (low communication cost), hierarchical intra-group and inter-group aggregation serves as a promising approach to tackle the non-IID issue in FedLoRA.
Hierarchical intra-group and inter-group aggregation serves as a promising approach to tackle the non-IID issue in FedLoRA.
% , which integrates naturally with FedLoRA due to the low communication cost \cite{zhang2022federated}
Motivated by these insights, we propose a novel hierarchical aggregation framework for FedLoRA, called \oursys, to address the challenges of resource constraints and data heterogeneity. 
Our unique findings include:
(1) Increasing the frequency of intra-group aggregation (called \textit{aggregation frequency}) significantly improves the convergence rate and final accuracy (Section \ref{ob-frequency}).
(2) Customizing the number of continuous fine-tuning layers close to the output (called \textit{depth}) for different groups helps to reduce the fine-tuning overhead while maintaining satisfactory fine-tuning performance (Section \ref{ob-depth}). 

% Based on these findings, \oursys aims to build an effective FedLoRA system.
Based on these findings, \oursys carefully organizes the device into groups with near-IID data distribution.
% Based on these findings, \oursys initiates by organizing the device into groups with near-IID data distribution.
In each round, the devices of each group fine-tune the model with designated depth to mitigate resource constraints and perform multiple intra-group aggregations before global aggregation to address non-IID data issues.
The difficulty of the system design lies in the interactions between aggregation frequency and depth.
Due to the limited resources, allocating high aggregation frequency compromises the applicable depth, whereas smaller depths tend to degrade the fine-tuning performance.
% On the one hand, a high aggregation frequency coupled with a large depth yields superior fine-tuning performance but incurs substantial resource consumption, resulting in slow convergence.
% On the other hand, a small depth with a low frequency reduces resource costs but results in suboptimal fine-tuning performance or, in some cases, failure to converge.
Thus, it is both necessary and challenging to simultaneously determine the optimal frequency and depth for heterogeneous groups, so as to strike an effective balance between fine-tuning performance and resource costs.
The main contributions of this paper are as follows:
% \vspace{-0.1cm}
\begin{itemize}
    \item We propose a hierarchical FedLoRA framework, called \oursys, to address data heterogeneity and resource constraints through an effective combination of aggregation frequency and depth adaptation.
    \item We analyze the joint influence of aggregation frequency and depth and obtain their coupled relationship. Then, we develop an efficient algorithm to balance the trade-off between fine-tuning efficiency and model accuracy.
    % \item We analyze the joint influence of aggregation frequency and depth and obtain their coupled relationship to determine the appropriate frequency and depth, aiming to balance the trade-off between fine-tuning efficiency and model performance.
    % \item We explore to jointly optimize aggregation frequency and depth upon their coupled relationship to better enhance the performance of FedLoRA.
    \item Comprehensive experiments on the physical platform show that \oursys enhances model accuracy by 1.6–4.2\% and accelerates fine-tuning by at least 2.1$\times$ compared to baselines.
    % \item The performance of \oursys is evaluated through a physical platform with 80 Jetson devices.
    % The results show that \oursys improves the final model accuracy by 1.6\% to 4.2\%, speeding up the fine-tuning process by at least 2.1$\times$, compared to the baselines.
\end{itemize}
% \vspace{-0.3cm}

% The rest of the paper is organized as follows.

%-------------------------------------------------------------------------------

%-------------------------------------------------------------------------------
\section{Background and Motivation}\label{sec:prelim}
\begin{figure*}[t]
    \centering
    \includegraphics[width=1.0\textwidth]{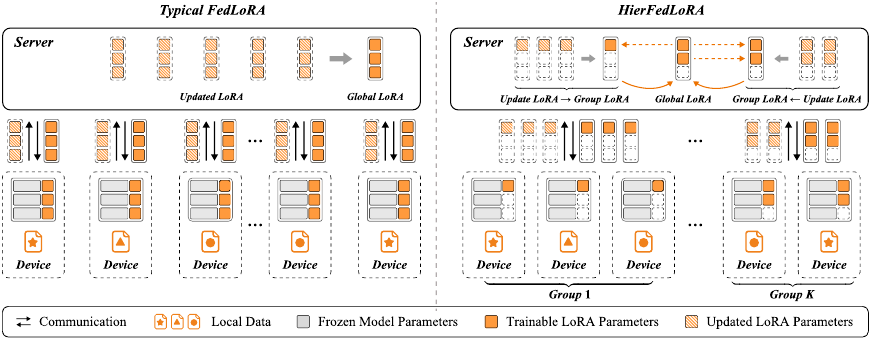}
    \vspace{-0.8cm}
    \caption{Illustration of typical FedLoRA and HierFedLoRA. Typical FedLoRA (left) applies a uniform fine-tuning strategy across all devices with periodic global aggregation at the PS; HierFedLoRA (right) partitions the devices into multiple near-IID groups, each adopting an appropriate fine-tuning depth and performing multiple intra-group aggregations before global aggregation.}
    \label{fig:illustration}
    % \vspace{-0.3cm}
\end{figure*}

\subsection{Federated Fine-Tuning with LoRA}
% \textbf{Low-Rank Adaptation.}
% Parameter-efficient fine-tuning (PEFT) \cite{houlsby2019parameter, hu2021lora, li2021prefix, zaken2021bitfit, zhao2022reduce, zhang2023fedpetuning} seeks to keep most parameters of LLMs frozen and fine-tune only additional lightweight parameters or a fraction of the parameters for downstream tasks. 
Parameter-efficient fine-tuning (PEFT) methods \cite{houlsby2019parameter, hu2021lora, li2021prefix, zaken2021bitfit, zhao2022reduce, zhang2023fedpetuning} reduce resource costs by freezing the pre-trained LLM and fine-tuning only a small subset of parameters for downstream tasks.
% Low-rank adaptation (LoRA) \cite{hu2021lora} is one of the most popular PEFT techniques that can achieve competent performance by only fine-tuning a small number of LoRA parameters (typically less than 1\% \cite{hu2021lora}).
% The core of LoRA is to represent each weight update as two rank decomposition matrices with much smaller ranks.
% Low-rank adaptation (LoRA) \cite{hu2021lora} is one of the most popular PEFT techniques that can achieve competent performance by only fine-tuning less than 1\% parameters of the LLM \cite{hu2021lora}.
Low-rank adaptation (LoRA) \cite{hu2021lora} is one of the most popular PEFT techniques, achieving competent performance by fine-tuning less than 1\% of the model parameters.
The core of LoRA is to represent each weight update as two rank decomposition matrices with much smaller ranks.
% During fine-tuning, LoRA only needs to optimize these two low-rank decomposition matrices, while keeping the pre-trained weights frozen. 
Specifically, for a pre-trained weight matrix $\mathcal{M} \in \mathbb{R}^{m \times q}$ ($m$ and $q$ are the dimension sizes of $\mathcal{M}$), LoRA injects low-rank decomposition $\Delta \mathcal{M} = \mathcal{B}\mathcal{A}$ as the tunable parameters. 
Note that, $\mathcal{B} \in \mathbb{R}^{m \times r}$ and $\mathcal{A} \in \mathbb{R}^{r \times q}$ are separately the project-down matrix and the project-up matrix, where $r$ denotes the rank of LoRA and is much smaller than both $m$ and $q$.
Formally, for a linear layer $y = \mathcal{M}x$, LoRA modifies the forward propagation as
$y = \mathcal{M}x + \mathcal{B}\mathcal{A}x$,
where $x$ and $y$ are the input tensors and the output tensors, respectively. 

% \textbf{Federated Fine-Tuning with LoRA.}
% Considering a distributed system with a parameter server (PS) and a set of $n$ devices, federated fine-tuning with LoRA (FedLoRA) is proposed to fine-tune the LLMs through a loose federation of devices, as illustrated in Figure \ref{fig:illustration} (left plot).
Considering a distributed system with a parameter server (PS) and a set of $n$ devices, federated fine-tuning with LoRA (called \textit{FedLoRA}) is introduced to fine-tune LLMs through a loose federation of devices, as illustrated in Fig. \ref{fig:illustration} (left plot).
Given the pre-trained LLM $\overline{\bm{w}}$, the goal of FedLoRA is to find the optimal LoRA parameters $\widetilde{\bm{w}}$, minimizing the loss function $f(\bm{w})$ as follows:
\vspace{-0.2cm}
\begin{equation}\label{loss_func}
    \min_{\bm{w} = \{\widetilde{\bm{w}}, \overline{\bm{w}} \}} f(\bm{w}) \triangleq \frac{1}{n} \sum_{i = 0}^{n - 1} f_i(\bm{w}_i)
    \vspace{-0.1cm}
\end{equation}
where $\bm{w}$ denotes the LoRA-enhanced LLM for simplicity, $f_i(\bm{w}_i) = \frac{1}{|\mathbb{D}_i|} \sum_{\xi_i \in \mathbb{D}_i} \ell(\bm{w}_i; \xi_i)$ is the loss function of the local LLM $\bm{w}_i$ on device $i$, and $\ell(\bm{w}_i; \xi_i)$ represents the loss over data samples $\xi_i$ on local dataset $\mathbb{D}_i$. 

During local fine-tuning in round $h$, device $i$ iteratively updates the LoRA parameters through $T$ steps, completing one epoch of its local dataset \cite{lin2021fednlp,zhang2022federated,cai2022fedadapter}, to minimize the local objective function as:
% During the local fine-tuning in round $h$, device $i$ updates the LoRA parameters $T$ times for a epoch on its local dataset \cite{lin2021fednlp,zhang2022federated,cai2022fedadapter}), to minimize the local objective function as:
% ($T$ is typically a constant representing the number of local steps required to complete one pass over the local dataset 
\vspace{-0.1cm}
\begin{equation}
    \widetilde{\bm{w}}_i^h \triangleq \widetilde{\bm{w}}_i^{h - 1} - \eta \cdot \sum_{\tau = 0}^{T - 1}  \nabla f_i(\widetilde{\bm{w}}_{i,\tau}^{h-1})
    \vspace{-0.1cm}
\end{equation}
where $\eta$ is the learning rate and $\nabla f_i(\widetilde{\bm{w}}_{i,\tau}^{h-1})$ is the gradient of the loss for LoRA parameters $\widetilde{\bm{w}}_{i,\tau}^{h-1}$ at local step $\tau \in [0, T - 1]$ in round $h$.
After that, the devices send the updated LoRA parameters to the PS for global aggregation as: 
\vspace{-0.1cm}
\begin{equation}
    \label{global-aggregation}
   \widetilde{\bm{w}}^{h+1} \triangleq \frac{1}{n} \sum_{i =0}^{n - 1} \widetilde{\bm{w}}_i^h
   \vspace{-0.1cm}
\end{equation}
Then, the PS distributes the newly aggregated LoRA parameters to the devices and moves to the next fine-tuning round. 
In doing so, the global LoRA-enhanced LLM can acquire knowledge from the local data of different devices without leaking their data privacy \cite{zhang2023fedpetuning}.
However, fine-tuning LLMs in typical FedLoRA remains challenging, as it imposes a significant computation burden (\eg, computing power and memory footprint) on resource-constrained devices \cite{kaddour2023challenges} and is susceptible to non-IID data \cite{zhang2022federated}, leading to degraded fine-tuning performance.

\subsection{Our Proposed Framework}
Herein, we propose \oursys, a hierarchical aggregation framework for FedLoRA, as illustrated in Fig. \ref{fig:illustration} (right plot). 
We draw on the extensively explored grouping methods \cite{liu2020client, wang2020towards, lee2020accurate, he2022improving}, and integrate hierarchical aggregation into FedLoRA.
\oursys partitions $n$ devices into $K$ groups using the proposed greedy algorithm (Section \ref{sec:device-group}), ensuring that the data distribution within each group approximates IID.
Each group $k$ consists of $n_k$ devices in round $h$, which are assigned to fine-tune $d_k^h$ continuous transformer layers (called depth) close to the output of the model and perform $\rho_k^h$ intra-group aggregations (called aggregation frequency) before global aggregation.
In the global round $h$, the update of LoRA parameters $\widetilde{\bm{w}}_{i, \tau}^{k,h}$ on device $i$ of group $k$ at local fine-tuning step $\tau$ ($\in [0, T - 1]$) can be expressed as follows:
\vspace{-0.1cm}
\begin{equation}
    \widetilde{\bm{w}}_{i, \tau}^{k,h} \triangleq \widetilde{\bm{w}}_{i, \tau - 1}^{k,h} - \eta \cdot \nabla f_i^k(\widetilde{\bm{w}}_{i, \tau - 1}^{k,h})
    % \vspace{-0.1cm}
\end{equation}
Once finishing updating the LoRA parameters $\hat{T} = T \slash \rho_k^h$ times, the devices of group $k$ upload the LoRA parameters to the PS for group aggregation as follows:
\vspace{-0.1cm}
\begin{equation}
    \widetilde{\bm{w}}_{\cdot, \tau}^{k,h} \triangleq \frac{1}{n_k} \sum_{i = 0}^{n_k - 1} \widetilde{\bm{w}}_{i, \tau}^{k,h}
    \vspace{-0.1cm}
\end{equation}
When each group $k$ finishes $\rho_k^h$ times global aggregation, the PS performs adaptive global aggregation by aggregating the $K$ group LoRA parameters.
% Formally, the aggregation of LoRA parameters in $l$-th ($\in [1, L]$) layer can be expressed as:
% \begin{equation}
%     \widetilde{\bm{w}}^{h}(l) \triangleq \frac{1}{n_l} \sum_{k = 0}^{n_l - 1} \widetilde{\bm{w}}^{k,h}(l)
% \end{equation}
% where $n_l$ denotes the number of $l$-th layers.
% Then, the objective of \oursys can be formulated by extending the loss function of equation (\ref{loss_func}) as follows:
% \begin{equation}
%     \min_{\bm{w} = \{\widetilde{\bm{w}}, \overline{\bm{w}} \}} f(\bm{w}) \triangleq \frac{1}{K} \sum_{k = 0}^{K - 1} f^k(\bm{w}^k)
% \end{equation}
% where $f^k(\bm{w}^k) \triangleq \frac{1}{n_k} \sum_{j = 0}^{n_k - 1} f_j^k(\bm{w}^k_j)$

% It is emergent to develop an efficient group-based FedFT system that considers both device resources and data distribution to tackle system and data heterogeneity.

% \begin{figure}[t]
%     \centering
%     \includegraphics[width=1.0\columnwidth]{figs/hierarchical-aggregation.pdf}
%     \vspace{-0.7cm}
%     \caption{The effect of hierarchical aggregation.}
%     \label{illu}
%     \vspace{-0.4cm}
% \end{figure}

\begin{figure}[t]
    \centering
    \subfigure[Accuracy v.s. time]{
        \vspace{-1cm}
        \begin{minipage}[t]{0.5\linewidth}
        \centering
        \includegraphics[width=1.7in]{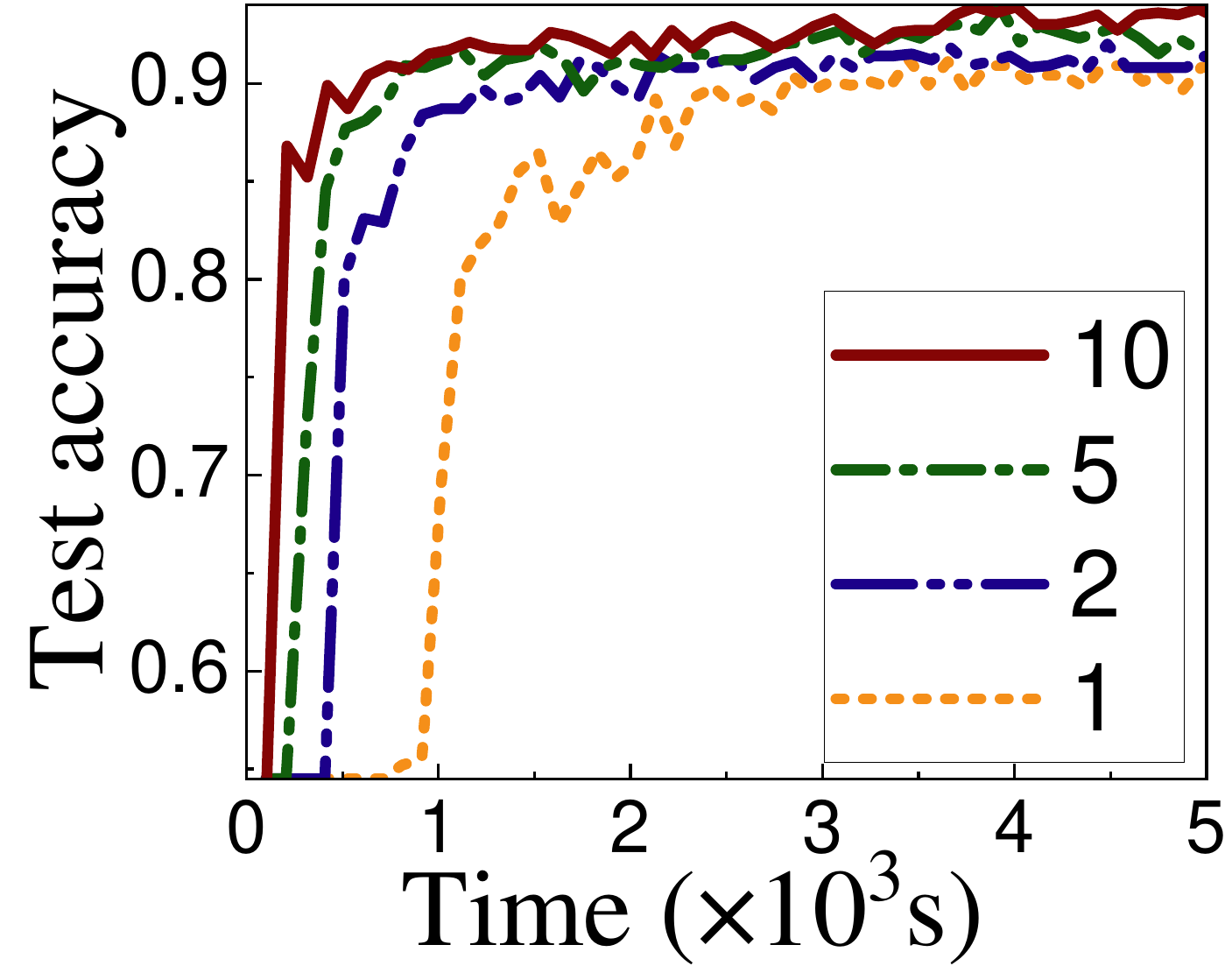}
        \end{minipage}%
        \vspace{-1cm}
        \label{1a}
    }%
    \subfigure[Resources costs to reach 90\%]{
        \begin{minipage}[t]{0.5\linewidth}
        \centering
        \includegraphics[width=1.7in]{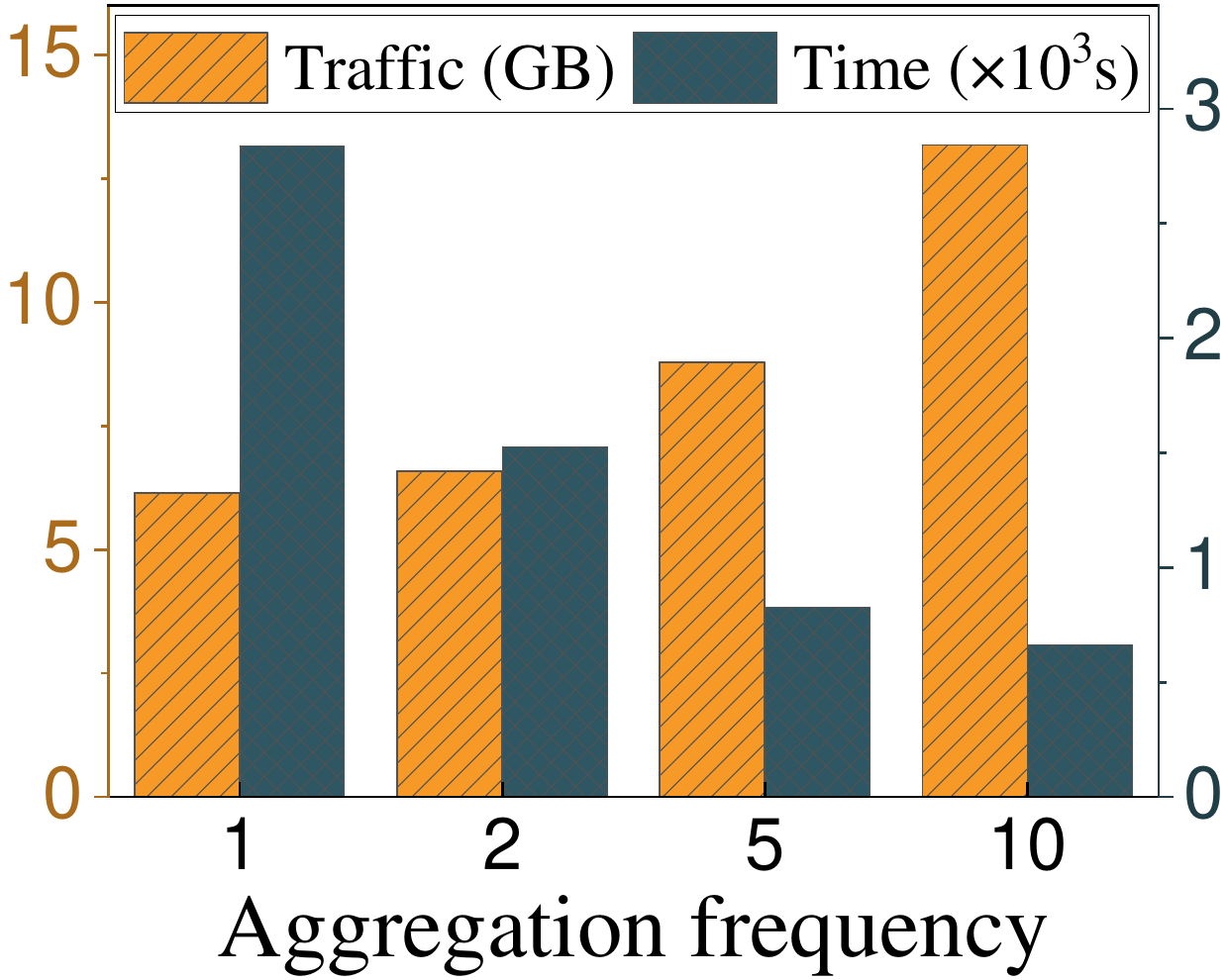}
        \end{minipage}%
        \label{1b}
    }%
    \centering
    \vspace{-0.5cm}
    \caption{Impact of aggregation frequency.}
    \label{ob-2}
    \vspace{-0.4cm}
\end{figure}

\subsection{Impact of Aggregation Frequency}\label{ob-frequency}
% The data distributions across different devices are inherently varied, resulting in diminished convergence rates and final accuracy in FedLoRA \cite{lin2021fednlp, babakniya2023slora}. 
% If the data distribution of each group is close to IID, periodically performing intra-group aggregations results in the group model approximating the one derived from IID data and thus improving the convergence. 
% \rednote{Device grouping does not automatically result in the optimal fine-tuning strategy in \oursys.}
% \bluenote{
% Although device grouping mitigates data heterogeneity to some degree, it alone does not guarantee an optimal fine-tuning strategy.
% }
% \bluenote{
% While device grouping effectively mitigates parameter drift among groups, it merely achieves a local optimum rather than the global fine-tuning strategy.
% }
% \bluenote{
% Although device grouping addresses data heterogeneity to some degree, it merely achieves a local optimum rather than the global fine-tuning strategy.
% }
Device grouping provides an effective foundation for addressing data heterogeneity.
If the data distribution of each group is close to IID, increasing the aggregation frequency helps the group model approximate the one derived from IID data, thereby improving convergence. 
In particular, when intra-group aggregation is performed after each step of local fine-tuning, the overall fine-tuning process is equivalent to that of centralized fine-tuning and thus mitigates the data heterogeneity issues.
% However, high frequency incurs more communication costs, while low frequency may still suffer from non-IID data. 
% Thus, it is important yet challenging to determine a proper aggregation frequency for each group.
However, while increasing the aggregation frequency improves fine-tuning performance, it also raises communication costs. 
On the other hand, a lower frequency reduces communication overhead but may still be insufficient to address the non-IID data issue. 
Therefore, it is important yet challenging to determine a proper aggregation frequency for each group.

To evaluate the impact of aggregation frequency, we fine-tune RoBERTa \cite{liu2019roberta} on 100 devices using the highly non-IID SST-2 dataset (following Dirichlet distribution with $\alpha = 0.1$ \cite{lin2021fednlp}) partitioned into 10 groups, where each group approximates an IID distribution.
Then, we conduct experiments to fine-tune RoBERTa \cite{liu2019roberta} with different aggregation frequencies $\rho = 1, 2, 5, 10$ with all devices set to the same depth, \ie, $d = 12$. 
The experimental results illustrated in Fig. \ref{ob-2} show that as the frequency increases, the fine-tuning performance significantly improves, while the communication costs also increase.
% For example, when increasing the frequency $\rho$ from $1$ to $\rho=2,5,10$, the final accuracy improves by about 0.2\%, 1.9\%, and 2.1\%, respectively, while incurring communication cost by about 7\%, 42.8\%, and 114\%, respectively and reducing the completion time to achieve the target accuracy of 90\%.
For example, increasing the frequency $\rho$ from 1 to 2, 5, and 10 improves the final accuracy by about 0.2\%, 1.9\%, and 2.1\%, respectively, and reduces the completion time to reach a 90\% target accuracy. 
However, this comes at the cost of increased communication, with costs rising by approximately 7\%, 42.8\%, and 114\%, respectively.
% Benefiting from the parameter-efficient nature of LoRA, periodic group aggregation significantly enhances the fine-tuning efficiency and performance when dealing with non-IID data, incurring only a modest increase in communication cost in FedLoRA.
Benefiting from the parameter-efficient nature of LoRA, increasing the aggregation frequency only incurs several MB of network traffic (\ie, transmission latency of a few seconds) per round while significantly enhancing convergence rates.

Therefore, increasing the aggregation frequency is both effective and necessary for addressing non-IID issues and enhancing fine-tuning efficiency.
However, due to the limited communication resources, high frequency may result in excessive communication costs, whereas low frequency tends to slow down the convergence process. 
To address the impact of data heterogeneity and resource constraints, \oursys assigns proper aggregation frequency to each group. 
The groups with strong devices are assigned with high frequency, and \textit{vice versa}. 
% Thus, \oursys address the heterogeneity issues while accelerating the convergence.

\begin{figure}[t]
    \centering
    \subfigure[Accuracy v.s. round]{
        \begin{minipage}[t]{0.5\linewidth}
        \centering
        \includegraphics[width=1.7in]{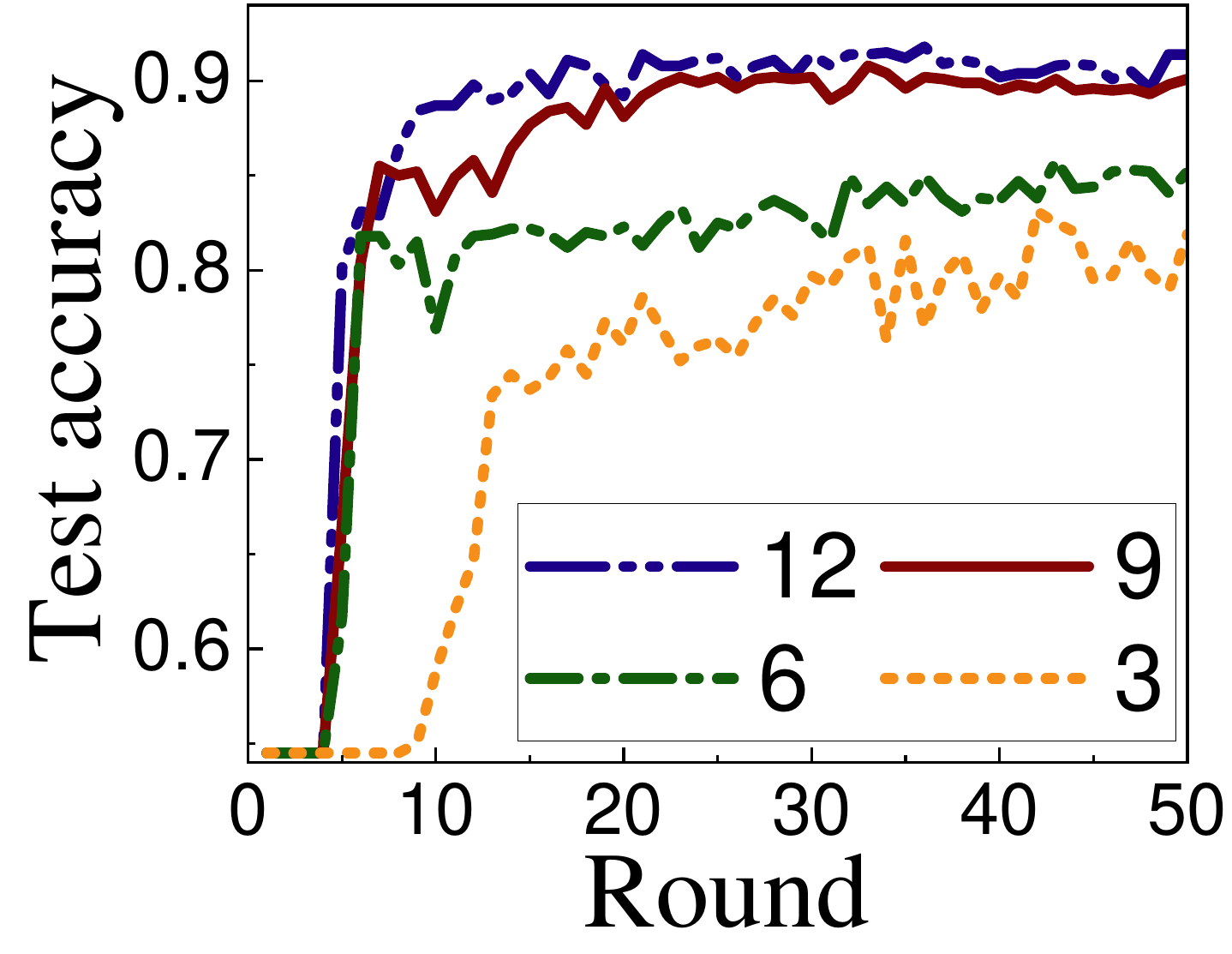}
        \end{minipage}%
        \label{3a}
    }%
    \subfigure[Resource costs per round]{
        \begin{minipage}[t]{0.5\linewidth}
        \centering
        \includegraphics[width=1.7in]{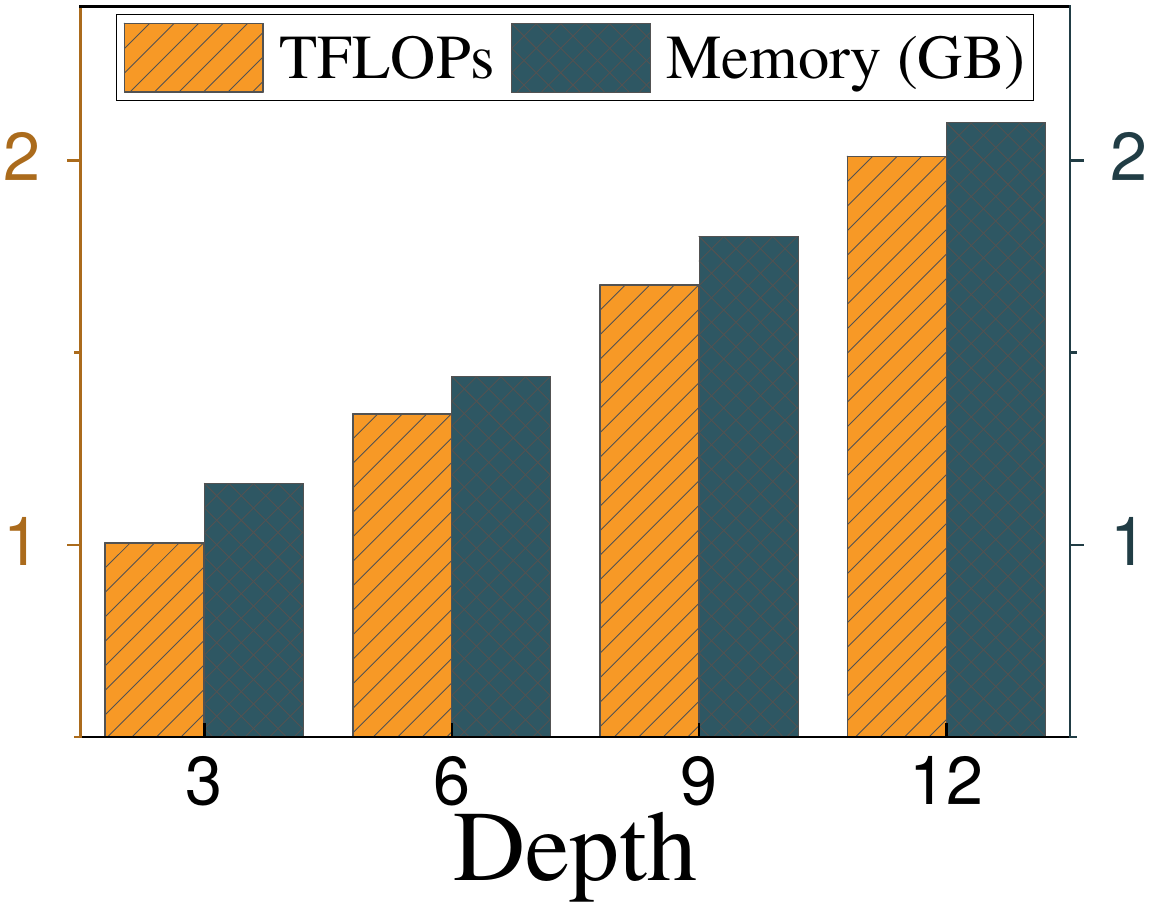}
        \end{minipage}%
        \label{3b}
    }%
    \centering
    \vspace{-0.4cm}
    \caption{Impact of fine-tuning depth.}
    \label{ob-1}
    \vspace{-0.5cm}
\end{figure}

\subsection{Impact of Fine-Tuning Depth}\label{ob-depth}
% Generally, PEFT methods (\eg, LoRA) still require complete forward and backward propagation through the entire model architecture, incurring computation costs comparable to full fine-tuning.
Existing works \cite{babakniya2023slora, yan2024federa, cho2023heterogeneous, bai2024federated, fang2024automated} can not effectively reduce the computation cost in FedLoRA, resulting in slow convergence.
Inspired by prior works \cite{brock2017freezeout, lee2019would, lin2021fednlp}, we explore the impact of LoRA fine-tuning depth, \ie, the number of tunable transformer layers from the output. 
% Generally, depth demonstrates a linear relationship with the activation memory and computation cost \cite{zhang2023lora}. 
Generally, depth demonstrates a linear relationship with resource costs in FedLoRA. 
A small depth restricts parameter updates and transmissions to those tunable layers, thereby reducing computation and communication costs.
% A small depth reduces memory usage by confining activation storage to the tunable layers and minimizes computation and communication costs by restricting parameter updates and transmissions to those layers.
However, small depths may hinder the model's adaptability to specific tasks, while large depths will increase the resource costs. 
Therefore, identifying the optimal LoRA depth is another critical factor for balancing fine-tuning performance and resource costs.

To demonstrate the impact of depth on fine-tuning performance, we also conduct a set of experiments with fine-tuning depths $d = 3, 6, 9, 12$ to demonstrate the impact of depth with all groups set to the sample aggregation frequency ($\rho = 1$). 
As illustrated in Fig. \ref{ob-1}, increasing the depth enhances the model's performance but leads to higher computation and communication costs (\eg, computing power, memory footprint, and network traffic).
% For example, when increasing the depth $d$ from 3 to 6, 9, and 12 enhances the final accuracy by approximately 2.7\%, 7.7\%, and 8.7\%, respectively, but also doubles the computing cost and memory usage.
For example, increasing the depth $d$ from 3 to 6, 9, and 12 improves the final accuracy by approximately 2.7\%, 7.7\%, and 8.7\%, respectively. 
However, this also increases computing power and memory footprint and linearly raises the communication cost.
% By the results, carefully determining LoRA depth is critical to improving fine-tuning performance, while saving computing and communication resources.
% To this end, \oursys partitions the devices into different groups based on their capabilities and data distributions. 
% \oursys assigns varying depths for each group to align with their capabilities. 
% Devices with strong computing and communication capabilities are assigned larger depths, while those with lower capabilities receive smaller depths, thereby reducing waiting times and further improving fine-tuning efficiency.

% 总结k和d各自的影响，讨论两者之间的相互关系是同时受到资源约束的，需要找到合适的权衡
\textbf{Discussion.}
% 通信是廉价的，计算是更大头的开销
According to the above insights, both aggregation frequency and depth significantly influence fine-tuning performance and resource costs.
However, due to the limited on-device resources, allocating high aggregation frequency compromises the applicable depth, whereas smaller depths tend to degrade the fine-tuning performance.
On one hand, high aggregation frequency combined with large depth effectively addresses non-IID challenges but introduces substantial computation and communication overhead, resulting in a slow convergence rate. 
Conversely, reducing both aggregation frequency and depth minimizes resource costs at the expense of fine-tuning performance or, in some cases, leads to convergence failure.
Therefore, in this paper, \oursys simultaneously determines the appropriate aggregation frequency and depth for each group, so as to balance the trade-off between resource costs and fine-tuning performance.
\section{System Design}\label{sec:algorithm}

\subsection{Overview}
\begin{figure}[t]
    \centering
    \includegraphics[width=1.0\columnwidth]{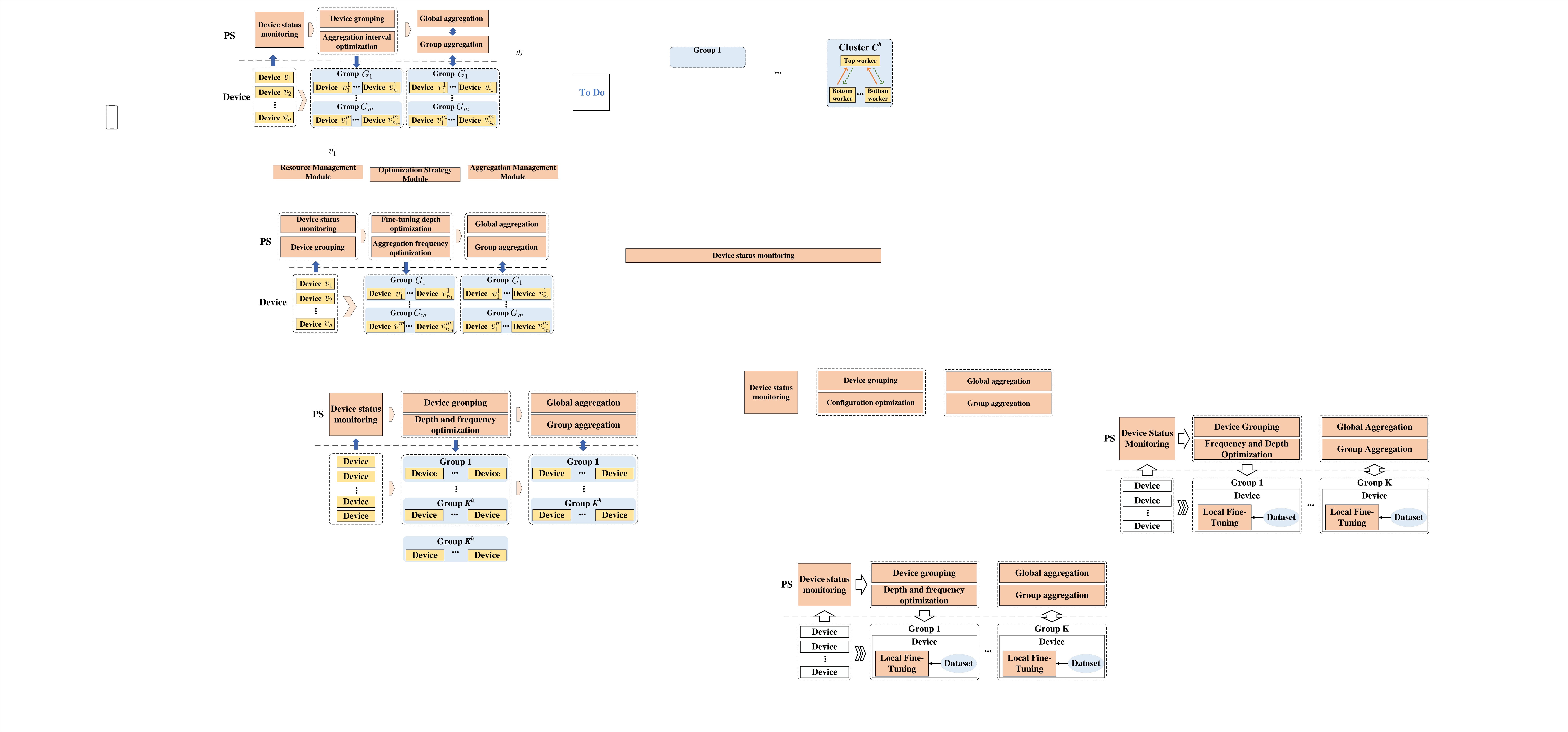}
    \vspace{-0.7cm}
    \caption{The proposed framework \oursys.}
    \label{framework}
    \vspace{-0.3cm}
\end{figure}

As illustrated in Figure \ref{framework}, \oursys consists of two key components with a total of six main modules, \ie, the PS with five modules and the device with one module.
The details of each module are as follows:

\textbf{Device Status Monitoring.}
The PS periodically collects information about the current working status of all devices (\eg, label distribution, computing, and communication capabilities) to make effective fine-tuning strategies. 

\textbf{Device Grouping.}
Based on the collected status information, the PS partitions the devices into proper groups using the proposed efficient grouping algorithm to promote fine-tuning efficiency.

\textbf{Frequency and Depth Optimization.}
In this module, the PS simultaneously determines the appropriate aggregation frequency and fine-tuning depth for each group.

\textbf{Local Fine-Tuning}
The device fine-tunes the LLM by updating LoRA parameters, periodically uploading them to the PS for aggregation, and reporting status (\eg, computing and communication time) for optimization.

\textbf{Group Aggregation.}
The module manages LoRA parameters for the groups, ensuring efficient organization and storage of updates, and performs intra-group aggregation.

\textbf{Global Aggregation.}
This module performs global aggregation, combining LoRA parameters across groups to derive the global model parameters.

\subsection{Device Status Monitoring}
In order to make an effective fine-tuning strategy, it is necessary to monitor the current status of all devices (\eg, label distribution, computing, and communication capabilities).
The collection of time-varying on-device resources (\eg, computing power and communication bandwidth) is necessary for device grouping and fine-tuning strategies optimization (\eg, frequency and depth optimization). 
Concretely, for device $i$ in round $h$, \oursys utilizes $\mu_i^h$ to denote the time required for local fine-tuning, which can be recorded by the devices directly, to denote the computing capacity.
Besides, since the uploading bandwidth is usually much smaller than the download bandwidth in typical WANs \cite{mcmahan2017communication, konecny2016federated}, we mainly focus on the uploading stage. 
Specifically, \oursys employs the uploading time $\beta_i^h$ of transmitting the LoRA parameters $\bm{w}_i^h$ from the device $i$ to the PS in round $h$ to indicate the communication capacity.
In global round $h$, PS collects recent computing time $\hat{\mu}_i^h$ and uploading time $\hat{\beta}_i^h$ from device $i$ and maintains the historical status.
Then, we introduce the moving average with the historical status of the devices to estimate the capacities of the device \cite{leroy2019federated}.
Accordingly, the PS estimates the computing time $\mu_i^h$ and the uploading time $\beta_i^h$ for device $i$ in round $h$ by calculating the moving average with $\alpha \in [0, 1]$ (\eg, $\alpha = 0.8$ in our experiments) as:
\begin{equation}
\mu_i^h = \alpha \cdot \mu_i^{h - 1} + (1 - \alpha) \cdot \hat{\mu}_i^h, \forall i \in [1,n], \forall h \in [1,H]
% \vspace{-0.1cm}
\end{equation}
\begin{equation}
\beta_i^h = \alpha \cdot \beta_i^{h - 1} + (1 - \alpha) \cdot \hat{\beta}_i^h, \forall i \in [1,n], \forall h \in [1,H]
\end{equation}
The primary focus of this work is not on improving status estimation techniques, and other advanced methods \cite{halperin2010predictable, yue2017linkforecast} can be easily integrated into \oursys.

To effectively handle the data heterogeneity, the local data distribution of each group together should be close to IID.
Herein, the label distribution, a vector $\Gamma = \{ \gamma_j \in [0, 1], j \in [1, C] \}$ ($\sum_{j = 1}^{C} \gamma_j = 1$) to parameterize a categorical distribution of class labels over $C$ classes \cite{zhao2018federated, liu2023yoga}, is utilized to guide the device grouping.

\subsection{Device Grouping}\label{sec:device-group}
Based on the collected status information of the devices, \oursys carefully partitions the devices into $K$ groups. 
Then, each group will be configured with a suitable aggregation frequency and fine-tuning depth by the PS.

To tackle the non-IID issue, the data distribution of each group needs to be close to IID.
We first define the IID distribution as $\Phi_0$.
If the data of all devices follows IID distribution, we can get $\Phi_0 = \frac{1}{n} \sum_{i = 1}^n \Gamma_i$, where $\Gamma_i$ is the label distribution of device $i$.
In group $k$, the label distribution of $n_k$ devices can be denoted as $\Phi_k = \frac{1}{n_k} \sum_{i = 1}^{n_k} \Gamma_i$.
Then, we utilize the KL divergence $KL(\Phi_k || \Phi_0)$ \cite{hershey2007approximating, goldberger2003efficient} to measure the gap between $\Phi_k$ and $\Phi_0$, which can be formulated as:
\vspace{-0.1cm}
\begin{equation}
    KL(\Phi_k || \Phi_0) = \sum_{c = 1}^C \Phi_k(\gamma_c) \log \frac{\Phi_k(\gamma_c)}{\Phi_0(\gamma_c)}
    \vspace{-0.2cm}
\end{equation}
To mitigate the negative impact of data heterogeneity, it is necessary to control the divergence $KL(\Phi_k || \Phi_0)$ as small as possible.
Additionally, in order to further improve the fine-tuning efficiency, we need to ensure that the completion times for all groups are approximately the same.
% Specifically, for each group $k$, there are $n_k$ devices. 
% In each global round, group $k$ performs $T$ rounds of local fine-tuning and periodically performs group aggregation $\rho_k^h$ times.
We can formulate the completion time $t_i^h$ of device $i$ in group $k$ as $t_{k,i}^{h} = \mu_{k,i}^{h} + \beta_{k,i}^{h}$.
Besides, the waiting time of device $i$ in group $k$ can be represented as $| t_{k,i}^{h} - t^{h}_k |$, where $t^{h}_k = \max\{ t_{k,i}^h \}_{i = 1}^{n_k}$ denotes the completion time of the slowest device of group $k$ in round $h$.
% The completion time of round $h$ can be denoted as $t^{h} = \max\{ t^{h}_k \}_{k = 1}^{K}$, which denotes the completion time of the slowest device in global in round $h$.
Thus, the average waiting time within group $k$ can be defined as:
\vspace{-0.1cm}
\begin{equation}\label{avg-waiting}
    \mathcal{W}^{h}_k = \frac{1}{n_k} \sum_{i = 1}^{n_k} | t_{k,i}^{h} - t^{h}_k |
    \vspace{-0.1cm}
\end{equation}
To minimize the waiting time of all devices in group $k$, we should group the $n_k$ devices, whose completion time of one round is close enough to each other, into the same cluster.

Considering data heterogeneity and resource constraints, it is challenging to partition these devices into appropriate groups.
We normalize the KL divergence $KL(\Phi_k || \Phi_0)$ and the average waiting time $\mathcal{W}_k^h$, and introduce a utility function to evaluate the effect of group $k$ in round $h$ as follows:
\begin{equation}
    \mathcal{U}_k^h = \lambda \cdot \mathcal{W}_k^h + (1 - \lambda) \cdot KL(\Phi_k || \Phi_0)
\end{equation}
where $\lambda$ is a weight coefficient used to balance $\mathcal{W}_k^h$ and $KL(\Phi_k || \Phi_0)$.
In round $h$, we need to partition all the devices carefully into appropriate groups to minimize $\sum_{k = 1}^K \mathcal{U}_k^h$, so that we can simultaneously address data heterogeneity and resource constraints, implementing efficient \oursys.

We propose a greedy algorithm to make an effective grouping strategy.
% Firstly, by the $K$-means algorithm (\eg, $K = n \slash 10$), according to the KL divergence of label distribution among devices, we divide the devices with small KL divergence into the same set and obtain $K$ set $S_1, S_2, ..., S_K$.
Firstly, by the $K$-means algorithm (\eg, $K = n \slash 10$), we divide the devices with small KL divergence into the same set and obtain $K$ set $S_1, S_2, ..., S_K$.
Next, we greedily construct the set $A$ including the device $j$ with maximum $t_j^h$ from each set $S_k$ and group devices into the group $k$ from set $A$ to make the $KL(\Phi_k || \Phi_0)$ smallest.
Subsequently, we repeat these operations to partition the remaining devices and create new groups, until all devices are partitioned into suitable groups.
Finally, we optimize the distribution of devices among groups through fine-grained adjustment, aiming to minimize the utility function $\sum_{k = 1}^{K}\mathcal{U}_k^h$.

\begin{algorithm}[t]
% \footnotesize
\caption{Device Grouping in Round $h$}\label{alg:device-group}

\KwIn{$B_i^h$, $\Gamma_i$, $\mu_i^h$, $\beta_i^h$, $K$.}
\KwOut{Device grouping strategy.}

Calculate $KL(\Phi_i || \Phi_j)$ for all devices $i$ and $j$, $i \neq j$;

Divide the devices into $K$ sets $S_1, S_2, \dots, S_K$ by calling the $K$-means algorithm.

Initialize $k = 1$;

\While{$S \neq \emptyset$}{
    Denote the distribution of group $k$ as $\Phi_k$;
    
    Select the device $j$ with the maximum $t_j^h$ from each $S_k$ to form set $A$;
    
    Select devices into the group $k$ from set $A$ to minimize $KL(\Phi_k || \Phi_0)$.
    
    $k \leftarrow k + 1$;
}
Minimize the utility function $\sum_{k = 1}^{K}\mathcal{U}_k^h$ by exchanging devices of different groups. 
\end{algorithm}

\subsection{Frequency and Depth Optimization}
\oursys relies on this module to determine the appropriate aggregation frequency $\rho_k^h$ and depth $d_k^h$ configurations for each group $k$ in round $h$, so as to address the challenges of data heterogeneity and resource constraints. 
% For simplicity, we use $\Psi = (\rho, d)$ to represent the combined configuration of aggregation frequency $\rho$ and fine-tuning depth $d$. 
Firstly, we use $\hat{u}$ to denote the computation cost for forward propagation of the entire LLM and represent the computation cost for backward propagation of a single transformer layer to update LoRA as $u$.
Then, the computation cost of configuration ($\rho_k^h$,$d_k^h$) can be formulated as follows:
\begin{equation}
    \mathcal{R}_{comp} (\rho_k^h,d_k^h) = \hat{u} + d_k^h \cdot u
\end{equation}
where $\hat{u}$ is the computation cost for forward propagation and u denotes the respective computation cost for backward propagation of a single transformer layer to update LoRA.
In addition, the communication cost of configuration ($\rho_k^h$,$d_k^h$) is formulated as:
\begin{equation}
    \mathcal{R}_{comm} (\rho_k^h,d_k^h) = \rho_{k}^h \cdot d_{k}^h \cdot b 
\end{equation}
where $b$ represents the communication consumption of LoRA parameters in a single transformer layer.
Assuming that the total computing and communication resource budgets of device $i$ in round $h$ are $\Pi_k^h$ and $\Omega_k^h$, respectively, the resource constraints can be expressed as follows:
\begin{equation}\label{constraint-comp}
    \mathcal{R}_{comp} (\rho_k^h,d_k^h) \leq \Pi_k^h
\end{equation}
\begin{equation}\label{constraint-comm}
    \mathcal{R}_{comm} (\rho_k^h,d_k^h) \leq \Omega_k^h
\end{equation}

Due to the complex and varying nature of federated environments, it is infeasible to predefine the optimal values of the combined configuration.
To this end, we attempt to learn relevant statistics online via the multi-armed bandit (MAB) theory, which has been extensively used to make sequential decisions in uncertain situations \cite{yoshida2020mab, ayache2023walk}.
The configuration decision problem can be naturally modeled as an MAB problem, where PS and the combined configuration (\ie, aggregation frequency and depth) correspond to the player and the arms, respectively.
In each round $h$, the PS decides which arm of the bandit is pulled.
After conducting fine-tuning based on $\rho_k^h$ and $d_k^h$, the player (\ie, PS) will observe the corresponding reward as follows:
\begin{equation}\label{reward-equation}
    \mathbb{R}(\rho_k^h,d_k^h) = \textbf{\textit{I}}(\frac{\Delta f_k^h}{\bar{\mathcal{R}}_k^h \cdot \mathcal{W}^{h}_k})
\end{equation}
where $\textbf{\textit{I}}(\cdot)$ is a normalization method that converts rewards into the range [0, 1]. 
$\Delta f_k^h = \frac{1}{n_k} \sum_{i=1}^{n_k} \Delta f_i^h$ represents the mean loss reduction for group $k$ during round $h$ and $\bar{\mathcal{R}}_k^h = v \cdot \mathcal{R}_{\text{comp}}(\rho_k^h,d_k^h) + (1 - v) \cdot \mathcal{R}_{\text{comm}}(\rho_k^h,d_k^h)$ denotes the normalized resource cost ($v$ is a weighted parameter) and $\mathcal{W}^{h}_k$ is the average waiting time of group $k$ in round $h$.
The rationale behind the reward function is to improve the convergence performance of FedLoRA in a resource-efficient way.

The objective of the MAB problem is to make sequential decisions to maximize the total reward obtained over a sequence of actions.
We extend the upper confidence bound (UCB) policy to address the MAB problem and introduce a resource-aware upper confidence bound (R-UCB).
Under the resource constraint, the R-UCB is designed to solve a bandit problem with a finite number of arms $\Psi = \{ \Psi_1, \Psi_2, ..., \Psi_m \}$, where each arm $\Psi_j = {(\rho_j, d_j)}$ corresponds to different combinations of aggregation frequency and depth, and $m = T \cdot L$ is the number of possible arms.
% The algorithm aims to maximize the upper bound of a confidence interval for the expected reward associated with each arm by adaptively partitioning the space of possible arms using a decision tree.
It employs an exploration and exploitation strategy to balance exploiting well-performed arms and exploring potential high-reward arms.
% Specifically, with a probability of $\epsilon$, a configuration is randomly selected to explore potentially better solutions, while with a probability of $1-\epsilon$, the current best configuration is chosen based on its reward values and selection history.
The exploitation and exploration are defined as:

1) \textbf{Exploitation.} 
Let $\mathcal{N}_h(\phi, \Psi_j^h) = \sum_{s=1}^{h-1} \phi^{h-s} \mathbbm{1}_{\{\Psi_k^s = \Psi_j^h\}}$ record the number of times that arm $\Psi_j^h$ is chosen, where $\phi$ is a discount factor and $\mathbbm{1}$ is indicator function.
$\mathbbm{1} = 1$ when $\Psi_k^s = \Psi_j^h$ and 0 otherwise.
The discounted empirical average is formulated as:
\begin{equation}
    \bar{\phi}_k(\phi, \Psi_j^h) = \frac{1}{\mathcal{N}_h(\phi, \Psi_j^h)} \sum_{s=1}^{h-1} \phi^{h-s} \mathbb{R}(\Psi_k^s) \mathbbm{1}_{\{\Psi_k^s = \Psi_j^h\}}
\end{equation} 

2) \textbf{Exploration.}
If the agent (\ie, PS) always selects the aggregation frequency and depth from the arm that is currently believed to be the best, it may miss the potential arm with a high reward.
To this end, R-UCB incorporates an exploration term into the upper bound.
Let $\hat{\mathcal{N}}_h(\phi) = \sum_{j=1}^m \mathcal{N}_h(\phi, \Psi_j^h)$ hold and the discounted padding function is defined as:
\begin{equation}
    \mathcal{P}_h(\phi, \Psi_j^h) = \sqrt{\frac{2\log \hat{\mathcal{N}}_h(\phi)}{\mathcal{N}_h(\phi, \Psi_j^h)}}
\end{equation}
The upper confidence bound in R-UCB is defined as:
\begin{equation}\label{eq-ucb}
    U_h(\Psi_j^h) = \bar{\phi}_k(\phi, \Psi_j^h) + \mathcal{P}_h(\phi, \Psi_j^h)
\end{equation}

We then choose the arm with the largest upper confidence bound.
The exploitation component computes a discounted weighted average of historical rewards, prioritizing more recent data through a decay factor $\phi$, while the exploration component grows proportionally with the duration since an arm was last selected.
By repeating the trial-and-error procedure, the player learns the decision strategy of fine-tuning to increase the reward in sequential actions.
The performance of the arm-pulling policy is evaluated by regret, defined as the difference between the expected reward from selecting the optimal arm $\Psi_{k^*}^h$ and the reward obtained by the actual one $\Psi_k^h$.
The goal of the MAB problem is to minimize the cumulative regret over $H$ rounds

\centerline{$\min \sum\limits_{h=1}^{H} \mathbb{E} [\mathbb{R}(\rho_{k^*}^h,d_{k^*}^h) - \mathbb{R}(\rho_k^h,d_k^h)]$}
% \vspace{-0.5cm}
\begin{equation}\label{problem}
    s.t.
    \begin{cases}
        \mathcal{R}_{comp} (\rho_k^h,d_k^h) \leq \Pi_k^h, & \forall k \in [K]\\
        \mathcal{R}_{comm} (\rho_k^h,d_k^h) \leq \Omega_k^h, & \forall k \in [K]
    \end{cases}
    % \vspace{-0.1cm}
\end{equation}
% \vspace{-0.4cm}

\begin{algorithm}[t]
\caption{Resource-aware Upper Confidence Bound for Group $k$}\label{two-layer-algo}
\SetKwFunction{process}{ProcessData}
\SetKwFunction{check}{CheckCondition}
\SetKwProg{Fn}{Function}{:}{}

\For{each round $h \gets 1$ \textbf{to} $H$}{
    \For{each configuration $\Psi_j \in \{\Psi_1, \Psi_2, ..., \Psi_m\}$}{
        Calculate the upper bound confidence bound according to Eq. \eqref{eq-ucb};
    }
    Choose the arm $\Psi_j^h$ with the largest upper confidence bound that satisfies the resource constraints of Eqs. \eqref{constraint-comp} and \eqref{constraint-comm};
        
    Conduct the fine-tuning procedure based on $\Psi_j^h$;

    Record the average waiting time based on Eq. \eqref{avg-waiting};
    
    Observe the actual reward according to Eq. \eqref{reward-equation};
}
\end{algorithm}

\subsection{Local Fine-Tuning}
In round $h$, device $i$ of group $k$ receives the latest LoRA parameters $\widetilde{\bm{w}}^{h,k}$ and the fine-tuning strategies (\ie, frequency $\rho_k^h$ and depth $d_k^h$) from the PS.
The device $i$ replaces the LoRA parameters with the received parameters according to the depth $d_k$.
Then, the device $i$ fine-tunes the LLM by updating the tunable LoRA parameters on its local dataset $\mathbb{D}_i$.
During the process of local fine-tuning in round $h$, device $i$ of group $k$ is associated with the local loss function $f(\bm{w}_i^{h,k})$, where $\bm{w} = \{\widetilde{\bm{w}}_i^{h,k}, \overline{\bm{w}} \}$ is the local model.
The loss of device $i$ on its local dataset $\mathbb{D}_i$ in round $h$ can be expressed as:
\begin{equation}
    f_i(\bm{w}_i^{h,k}) = \frac{1}{| \mathbb{D}_i |} \sum \ell (\bm{w}_i^{h,k}; \xi_i)
\end{equation}
where $\xi_i$ is a batch of data samples in $\mathbb{D}_i$, and $\ell (\bm{w}_i^{h,k}; \xi_i)$ is the local loss over data $\xi_i$.
In general, the device utilizes a stochastic gradient descent algorithm, \eg, Adam \cite{kingma2014adam} or AdamW \cite{loshchilov2017decoupled}, to iteratively update the LoRA parameters based on the gradient over each batch of data samples.
Formally, the process of updating the LoRA parameters $\widetilde{\bm{w}}_i^{h,k}$ at local fine-tuning step $\tau$ can be expressed as:
\begin{equation}
    \widetilde{\bm{w}}_{i,\tau}^{h,k} = \widetilde{\bm{w}}_{i,\tau - 1}^{h,k} - \eta \cdot \nabla f_i(\widetilde{\bm{w}}_{i,\tau - 1}^{h,k})
\end{equation}
where $\eta$ is the learning rate and $\nabla f_i(\widetilde{\bm{w}}_{i,\tau - 1}^{h,k})$ is the gradient of the loss for the LoRA parameters $\widetilde{\bm{w}}_{i,\tau - 1}^{h,k}$.

\subsection{Group Aggregation}
During local fine-tuning, the device $i$ periodically uploads the LoRA parameters to the PS for $\rho_k^h$ times of intra-group aggregation.
% After that, the device $i$ uploads the LoRA parameters and the status information to the PS for global aggregation and fine-tuning strategy optimization.
Benefiting from the parameter-efficient nature of LoRA, the server only needs to maintain the latest LoRA parameters for each group, introducing negligible storage and maintenance overhead.
Specifically, in round $h$, once completing $\tau$ times of local iterations, where $\tau \mod (T/\rho_k^h) = 0$, the 
device $i$ of group $k$ will send the LoRA parameters $\widetilde{\bm{w}}_{i,\tau - 1}^{h,k}$ to the PS for group aggregation as follows:
\vspace{-0.1cm}
\begin{equation}
    \widetilde{\bm{w}}_{:,\tau - 1}^{h,k} = \frac{1}{n_k} \sum_{i = 1}^{n_k} \widetilde{\bm{w}}_{i,\tau - 1}^{h,k}
    \vspace{-0.1cm}
\end{equation}
The aggregated parameters $\widetilde{\bm{w}}_{:,\tau - 1}^{h,k}$ will be immediately sent to the device for the following fine-tuning process.

\subsection{Global Aggregation}
When completing a total of $T$ times of local iteration and $\rho_k^h$ times of group aggregation, the device $i$ of group $k$ sends the latest LoRA parameters $\widetilde{\bm{w}}_{i,T}^{h,k}$ and the collected status information (\eg, computing and communication time) during the current round of local fine-tuning to the PS.
The PS then performs layer-wise aggregation for the received LoRA parameters from different groups \cite{ma2022layer, liu2023yoga, wei2024flexora}.
Specifically, the aggregation of tunable LoRA parameters in $l$-th transformer layer can be expressed as follows:
\vspace{-0.1cm}
\begin{equation}
    \widetilde{\bm{w}}^{h+1}(l) = \frac{1}{n_l} \sum_{i = 1}^{n_l} \widetilde{\bm{w}}_{i}^{h}(l)
    \vspace{-0.1cm}
\end{equation}
Once the aggregation is completed, the PS sends the aggregated parameters to all devices, along with the generated fine-tuning strategies for the next round.

\section{Evaluation}\label{sec:evaluation}
\begin{table}[!t]
    % \vspace{-0.35cm}
    \caption{Technical Overview of Jetson Devices}
    \vspace{-0.2cm}
    \centering
    \begin{tabular}{lcc}
        \Xhline{1pt}
        \textbf{Jetson} & \textbf{AI Performance} & \textbf{GPU Type} \\ 
        \Xhline{0.7pt}
        TX2 & 1.33 TFLOPS & 256-core Pascal \\ \hline
        NX & 21 TOPS & 384-core Volta\\ \hline
        AGX Xavier & 22 TOPS &   512-core Volta \\
        \Xhline{0.7pt}
        \textbf{Jetson} & \textbf{CPU Type} & \textbf{ROM} \\ 
        \Xhline{0.7pt}
        TX2 & Denver 2 and ARM 4 & 8 GB LPDDR4\\ \hline
        NX & 6-core Carmel ARM 8 & 8 GB LPDDR4x\\ \hline
        AGX Xavier& 8-core Carmel ARM 8 & 32 GB LPDDR4x \\
        \Xhline{1pt}
        % \textbf{Jetson}& \textbf{CPU Frequency} & \textbf{GPU Frequency} \\
        % \Xhline{0.7pt}
        % TX2 & 1.2GHz & 0.85Ghz\\ \hline
        % NX & 1.2GHz & 0.8Ghz\\ \hline
        % AGX Xavier& 1.45GHz & 0.9Ghz\\ 
        % \Xhline{1pt}
    \end{tabular}
    \label{jetson-info}
    % \vspace{-0.3cm}
\end{table}

\subsection{Experimental Settings}
\label{evaluation}
\textbf{System Implementation.}
Extensive experiments are conducted on a prototype system with one PS and 80 devices to evaluate the performance of \oursys.
Specifically, we employ a deep learning GPU workstation as the PS, which is equipped with an Intel(R) Core(TM) i9-10900X CPU, four NVIDIA GeForce RTX 2080Ti GPUs, and 256 GB RAM.
In addition, we specify 80 NVIDIA commercial developer kits, including 30 Jetson TX2 kits, 40 Jetson NX kits, and 10 Jetson AGX kits, as devices to construct the heterogeneous system. 
The detailed technical specifications of Jetson TX2, NX, and AGX kits are listed in Table \ref{jetson-info}.

The software platform is built based on Docker Swarm \cite{merkel2014docker, naik2016building}, a distributed software development kit that helps build distributed systems with the ability to monitor the status of each device, and PyTorch \cite{paszke2019pytorch}, a deep learning library to facilitate the implementation of model training on devices. 
In addition, we adopt MPI (Message Passing Interface) \cite{gabriel2004open}, which includes a collection of sending and receiving functions, to streamline communication between the PS and devices.

\textbf{Settings of System Heterogeneity.}
To emulate the heterogeneous computing and communication capabilities among devices, we present the following setups.

1) \textbf{\textit{For Computing.}}
By specifying different modes of the Jetson devices (\ie, Jetson TX2, NX, and AGX), our prototype system enables these devices to work with varying computing capabilities.
Specifically, Jetson TX2 offers four configurable modes, whereas the Jetson NX and AGX support up to eight modes.
The Jetson AGX with the mode 0 (\ie, the highest performance mode of Jetson AGX) achieves training by 100$\times$ faster than the TX2 with the mode 1 (\ie, the lowest performance mode of Jetson TX2).
Besides, to reflect resource varying over time, the devices are configured to randomly change the mode every 20 rounds.

2) \textbf{\textit{For Communication.}}
To replicate the practical network environment, all devices are connected to the PS via Wi-Fi routers in the prototype system.
Concretely, the devices are randomly shuffled and divided into four groups, with each group containing 20 devices. 
Then, these groups are placed at different locations, \ie, 2m, 8m, 14m, and 20m, away from the Wi-Fi routers.
Due to random channel noise and competition among devices, the bandwidth between the PS and devices varies dynamically during the training.
The bandwidth of devices is measured by iperf3 \cite{tirumala1999iperf}, which fluctuates between 1Mb/s and 30Mb/s.

\begin{table}[t]
\renewcommand{\arraystretch}{1.1}
    \centering
    \caption{Overview of the Evaluation Tasks}
    \vspace{-0.2cm}
    % \vspace{-5px}
    % \resizebox{\textwidth}{!}{
    \begin{tabular}{lccccccc}
         \Xhline{1pt}
         \textbf{Task} &  \textbf{Dataset} & \makecell{\textbf{\# Training}} &\makecell{\textbf{\# Test}} \\
         \Xhline{0.7pt}
         Sentiment Analysis  & SST-2 & 67,349   & 1,821  \\ \hline
         Question Answering   & QNLI & 104,743 & 5,463  \\ \hline
         Semantic Equivalence  & QQP & 363,846  & 40,430  \\ \hline
         Textual Entailment  & MNLI & 392,702  & 9,815  \\
         % GSM-8K   & i.i.d. & 7473 & 1,319  \\ \hline
         % MMLU  & i.i.d. & 20,000  & 2,000 \\ \hline
         % Code Generation & Rosetta & 7954 & Prog. Lang. & HumanEvalX & 656 & Llama3.2-1B \\
         % Question Answering & Dolly & 15015 & 8 & Category &  Helm & NA \\
         % Public Dataset & Alpaca & 52002 & \multicolumn{3}{c}{--------------------------------------------------------} & Llama3.2-1B \\
         \Xhline{1pt}
    \end{tabular}
    % }
    \label{tab:datasets}
    % \vspace{-0.6cm}
\end{table}

\textbf{Tasks and Models.} 
% We evaluate the performance of \oursys on four representative natural language understanding applications with two transformer-based models.
To emulate real-world scenarios, we conduct extensive experiments on four representative tasks in mobile applications, \ie, sentiment analysis, question answering, semantic equivalence, and textual entailment. 
Following the previous works \cite{lin2021fednlp, zhang2022federated, babakniya2023slora, cai2022fedadapter, 298559}, we employ two well-suited LLMs for these tasks, \ie, RoBERTa \cite{liu2019roberta} and DeBERTa-large \cite{he2021debertav3}.
% To simulate real-world scenarios, we conduct sets of experiments on four representative tasks in mobile applications, \ie, sentiment analysis, question answering, semantic equivalence and textual entailment, on two LLMs, \RoBERTa \cite{liu2019roberta} and DeBERTa-large \cite{he2021debertav3}, which are particularly well-suited for such applications \cite{zhang2022federated, 298559}. 
% on four general language understanding applications using two encoder-only models, which are particularly well-suited for such applications \cite{zhang2022federated, 298559}. 
% These experiments align with our focus on labeled data heterogeneity.

1) \textbf{\textit{Sentiment Analysis}} aims to extract subjective information from text data such as positive, negative, or neutral \cite{medhat2014sentiment}.
The Stanford Sentiment Treebank (SST-2) dataset \cite{wang2018glue}, which consists of 70,042 sentences (67,349 training samples and 1,821 test samples) from movie reviews with human sentiment annotations, is adopted for sentiment analysis. 
% We train tailored BERT \cite{devlin2018bert} (with 12 transformer blocks and 110M parameters) for pre-experiments and 
We fine-tune the RoBERTa-base model \cite{liu2019roberta} with 125M parameters on SST-2.
% RoBERTa is composed of 12 transformer  layers, a 768 $\times$ 2 fully connected hidden layer, and a softmax output layer.
% We train tailored RoBERTa \cite{liu2019roberta} (with 12 transformer blocks and 124M parameters, which is an extension and refinement of BERT \cite{devlin2018bert}) on SST-2.

% 2) \textbf{\textit{Question Answering.}} 
2) \textbf{\textit{Question Answering}} focuses on generating concise and accurate answers to questions of natural language by comprehending the query and retrieving relevant information from data sources \cite{rajpurkar2016squad}.
We utilize the Question-answering Natural Language Inference (QNLI) dataset \cite{wang2018glue}, a classification dataset consisting of question-sentence pairs in the corresponding context, for this task. 
The dataset includes 104,743 and 5,463 samples for training and test, respectively.
% We employ the same RoBERTa model on SST-2 for QNLI dataset
% The same model as on SST-2 is employed on the QNLI dataset.
The foundation model employed on QNLI is the same as that on SST-2 (\ie, RoBERTa).

% We adopt RoBERTa with different output layer on xxx

% The objective of the dataset is to determine whether the context sentence contains the answer to the question.
% A tailored RoBERTa is trained on this dataset.

% 3) \textbf{\textit{Semantic Equivalence.}}
3) \textbf{\textit{Semantic Equivalence}} determines whether two given texts convey the same meaning or not, disregarding differences in word choice or syntax \cite{chandrasekaran2021evolution}. 
We use the Quora Question Paris (QQP) dataset \cite{wang2018glue} for evaluation.
The QQP dataset is a collection of question pairs composed of 363,846 and 40,430 samples for training and test, respectively, from the website Quora. 
DeBERTa-large \cite{he2021debertav3} with 430M parameters is fine-tuned on QQP.
% , consisting of 24 transformer layers, a 768 $\times$ 2 fully connected layer, and a softmax output layer.
% We train a customized DeBERTa-V3-Large model \cite{he2021debertav3} on QQP, which is the latest version of the DeBERTa model series. 
% The DeBERTa-V3-Large contconsisting of 24 transformer layers

% 4) \textbf{\textit{Textual Entailment.}}
4) \textbf{\textit{Textual Entailment}} reasons the logical relationship between a given premise and a corresponding hypothesis, which can be classified as entailment, contradiction, or neutral \cite{ghuge2014survey}. 
The Multi-Genre Natural Language Inference (MNLI) dataset \cite{wang2018glue}, a crowdsourced collection of sentence pairs with textual entailment annotations, is adopted for textual entailment.
The dataset consists of 392,702 and 9,815 samples for training and test, respectively.
% The objective of this task is to determine the logical relationship between a given premise and a corresponding hypothesis, which can be classified as entailment, contradiction, or neutral. 
Same as QQP, we fine-tune the DeBERTa-large for this task.

% \begin{table}
%     \caption{Datasets used in the experiments}
%     \centering
%     \resizebox{\linewidth}{!}{
%     \begin{tabular}{|c|c|c|c|c|}
%     \hline
%     Dataset & Application  & \# Train & \# Test & Model \\
%     \hline
%     SST-2    & Sentiment Analysis    & 67,349    & 1,821       & RoBERTa  \\ \hline
%     % MRPC     & Textual similarity      & 3,668     & 408       & F1 score \\
%     QNLI     & Question Answering     & 104,743   & 5,463     & RoBERTa \\ \hline
%     % STS-B    & Semantic similarity          & 5,749     & 1,500     & Pearson/Spearman correlation   \\
%     QQP      & Semantic Equivalence & 363,846   & 40,430    & DeBERTa-V3-Large \\ \hline
%     MNLI     & Textual Entailment & 392,702   & 9,815     & DeBERTa-V3-Large \\ 
%     % RTE      & Textual entailment         & 2,490     & 277       & Accuracy \\
%     % WNLI     & Coreference resolution     & 635       & 71        & Accuracy \\
%     % CoLA     & Linguistic acceptability   & 8,551     & 1,043     & Matthews correlation \\
%     % AGNEWS  & Text classification        & 120,000      & 7,600   & Accuracy \\
%     % 20NEWS  & Text classification       & 11,314        & 7,532   & Accuracy \\
%     \hline
%     \end{tabular}}
%     \label{tab-glue-datasets}
% \end{table}

\textbf{Setting of Data Heterogeneity.}
In the experiments, training samples of each device are drawn independently by a vector $\gamma$.
To create non-IID datasets, we draw from a Dirichlet distribution \cite{lin-etal-2022-fednlp}, \ie, $\gamma \sim Dir(\delta q)$, where $q$ characterizes a prior class distribution, and $\delta > 0$ is a concentration parameter controlling the identicalness among devices.
With $\delta \rightarrow \infty$, all devices have identical distributions to prior class distribution (\ie, IID); with $\delta \rightarrow 0$, each worker holds data samples from only one class, which indicates a high degree of data heterogeneity.
We specify 6 values (\eg, $\infty, 1, 0.5, 0.25, 0.2, 0.1$) for $\delta$ to generate different data distributions that cover a spectrum of identicalness, and define $p = 1 \slash \delta$ (\ie, $p = 0, 1, 2, 4, 5, 10$) to quantify the non-IID levels.
The degree of data heterogeneity increases as $p$ increases, and $p = 0$ is a special case of IID data distribution..

\textbf{Baselines.}
We adopt four approaches as baselines to evaluate the effectiveness of \oursys.

1) \textbf{BaseFedLoRA} 
integrates vanilla LoRA \cite{hu2021lora} into FedLLM, where all the devices fine-tune the same local model with the identical rank applied to all layers.

2) \textbf{FedDeRA} \cite{yan2024federa} improves BaseFedLoRA by applying SVD to pre-trained weights to initialize LoRA before the federated fine-tuning process.

3) \textbf{HetLoRA} \cite{cho2023heterogeneous} 
is an advanced LoRA-based approach for FedLLM, which assigns each device with a diverse but appropriate LoRA rank to perform fine-tuning.

4) \textbf{FlexLoRA} \cite{bai2024federated}
employs a dynamic parameter allocation strategy to adjust LoRA rank and utilizes SVD for weight redistribution.

\begin{figure*}
	\centering
	\subfigure[SST-2]{
		\begin{minipage}[b]{0.23\textwidth}
			\includegraphics[width=1\textwidth]{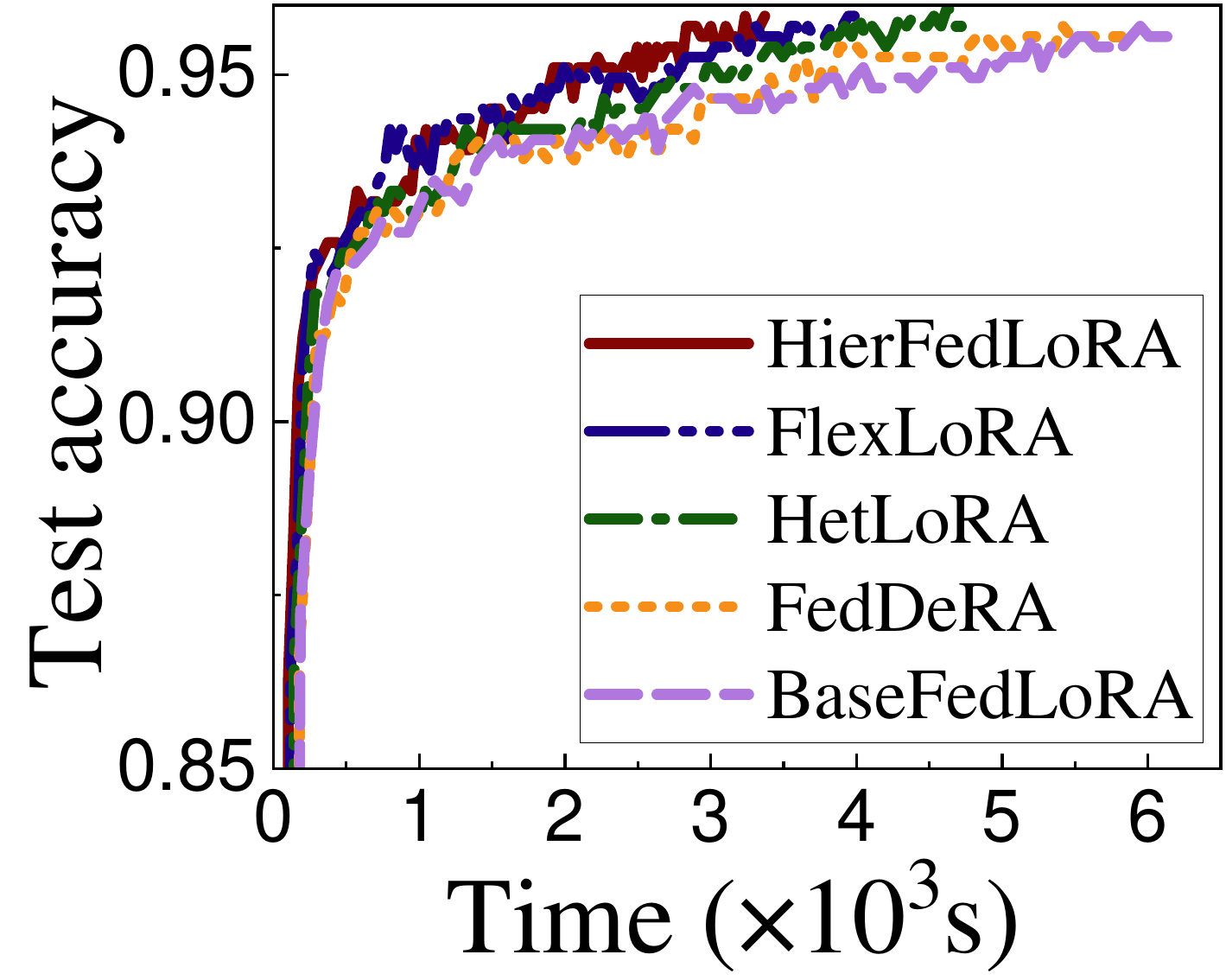} 
		\end{minipage}
		\label{time-acc-iid-sst2}
	}
    	\subfigure[QNLI]{
    		\begin{minipage}[b]{0.23\textwidth}
   		 	\includegraphics[width=1\textwidth]{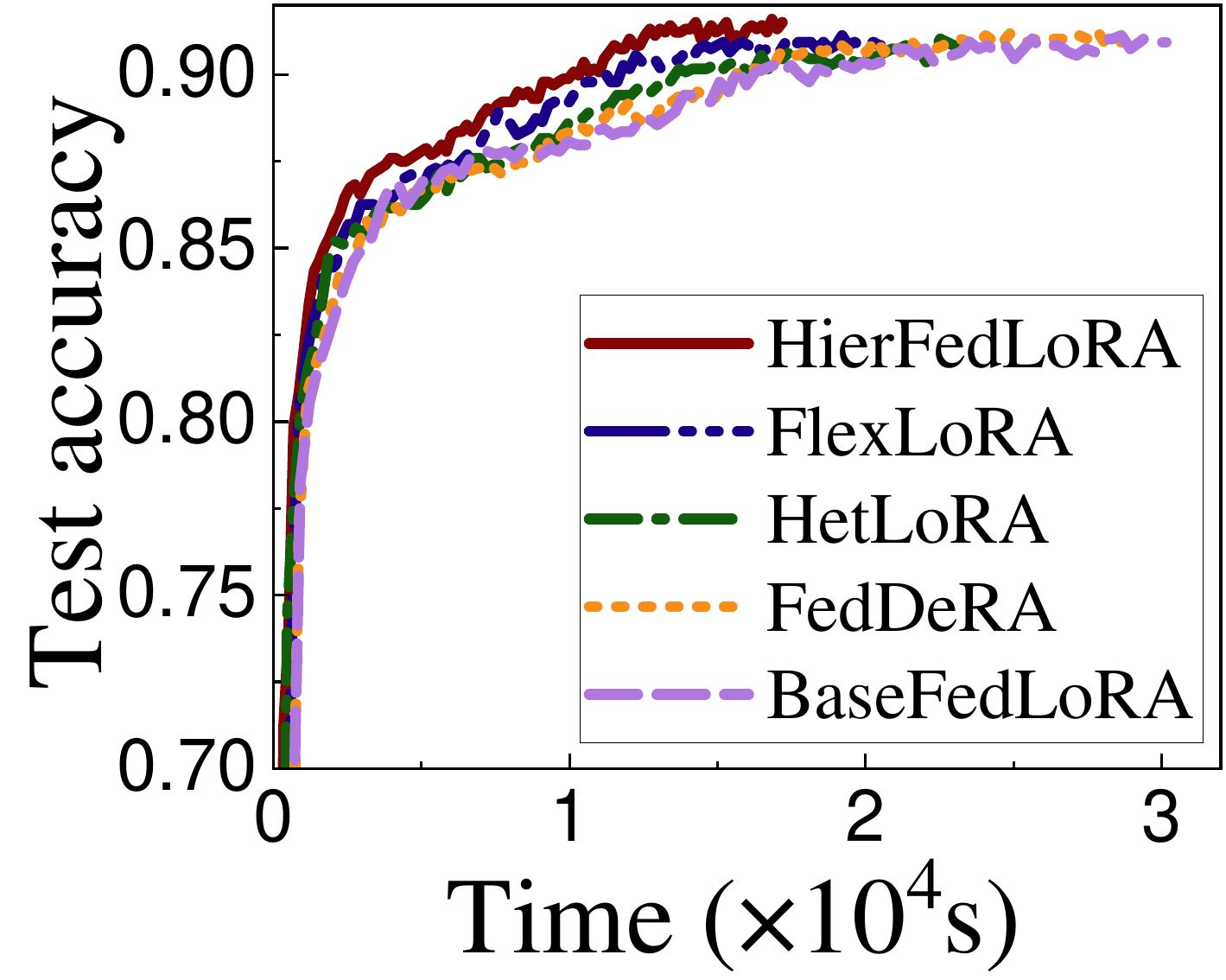}
    		\end{minipage}
		\label{time-acc-iid-qnli}
    	}
	% \\ 
	\subfigure[QQP]{
		\begin{minipage}[b]{0.23\textwidth}
			\includegraphics[width=1\textwidth]{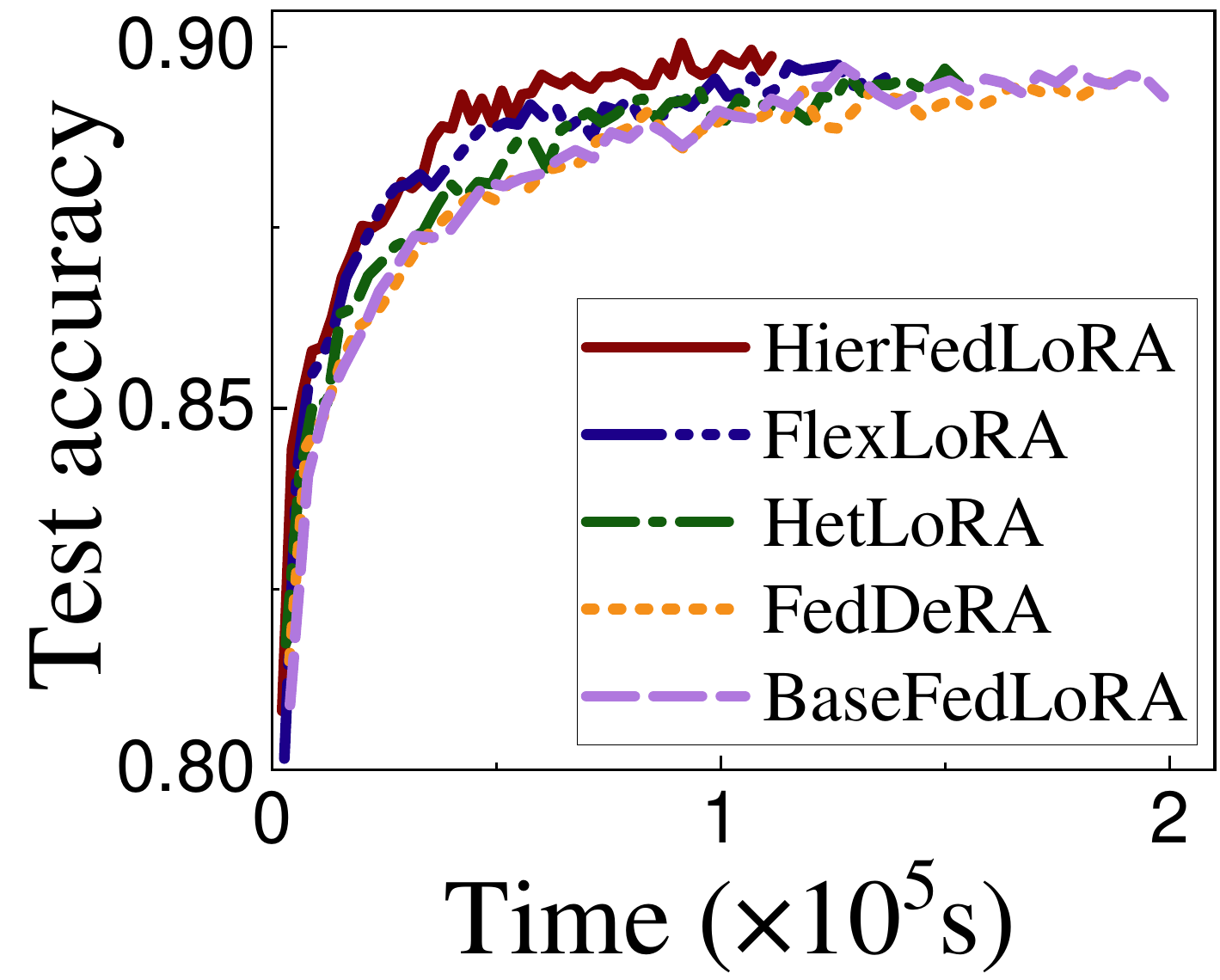}
		\end{minipage}
		\label{time-acc-iid-qqp}
	}
    	\subfigure[MNLI]{
    		\begin{minipage}[b]{0.23\textwidth}
		 	\includegraphics[width=1\textwidth]{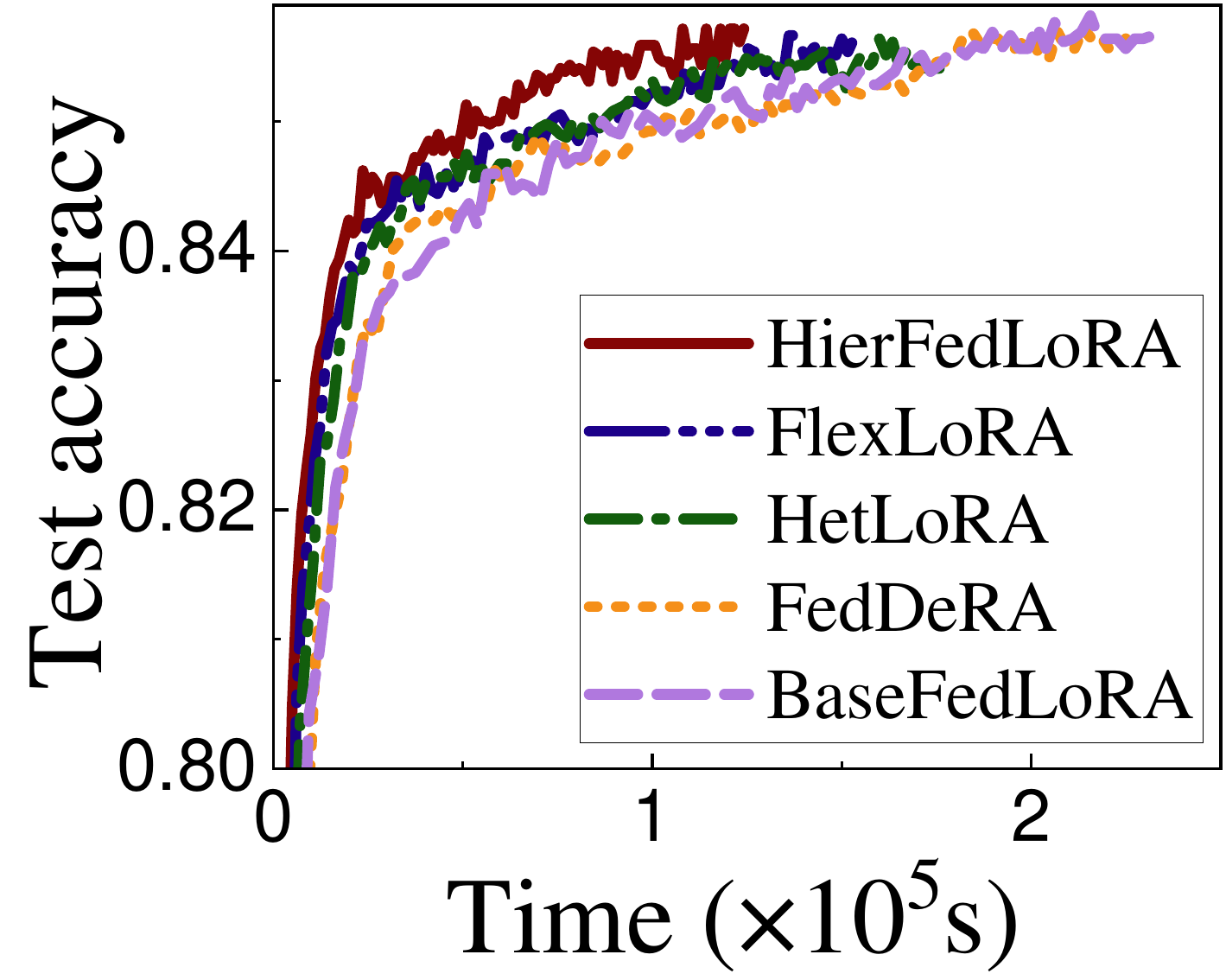}
    		\end{minipage}
		\label{time-acc-iid-mnli}
    	}
        \vspace{-0.4cm}
	\caption{Time-to-accuracy of five approaches on the four IID datasets.}
        \vspace{-0.3cm}
        \label{time-acc-iid}
\end{figure*}

\begin{figure*}
	\centering
    % \hspace{3.5mm}
	\subfigure[SST-2]{
		\begin{minipage}[b]{0.23\textwidth}
			\includegraphics[width=1\textwidth]{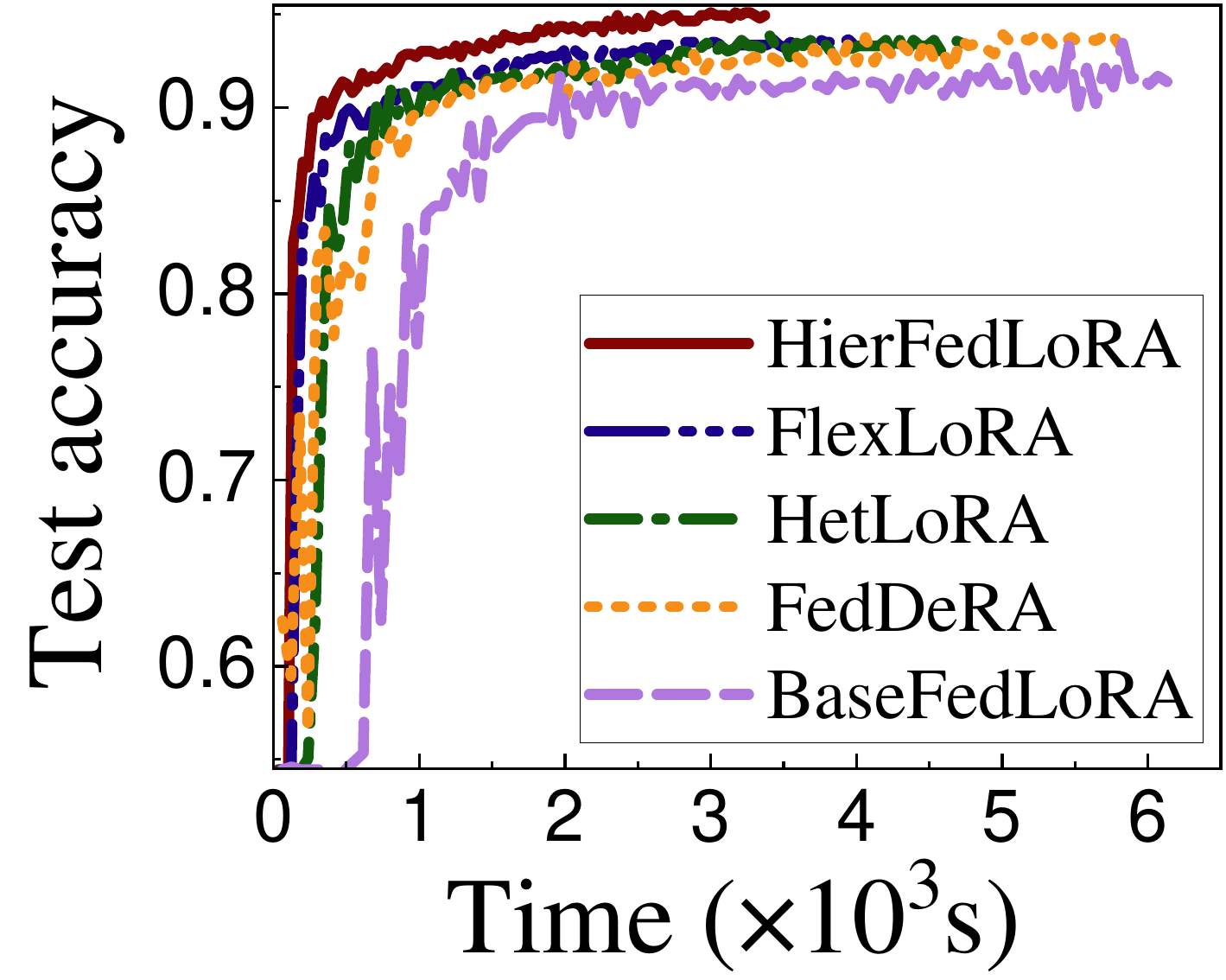}
		\end{minipage}
		\label{time-acc-non-iid-sst2}
	}
    	\subfigure[QNLI]{
    		\begin{minipage}[b]{0.23\textwidth}
   		 	\includegraphics[width=1\textwidth]{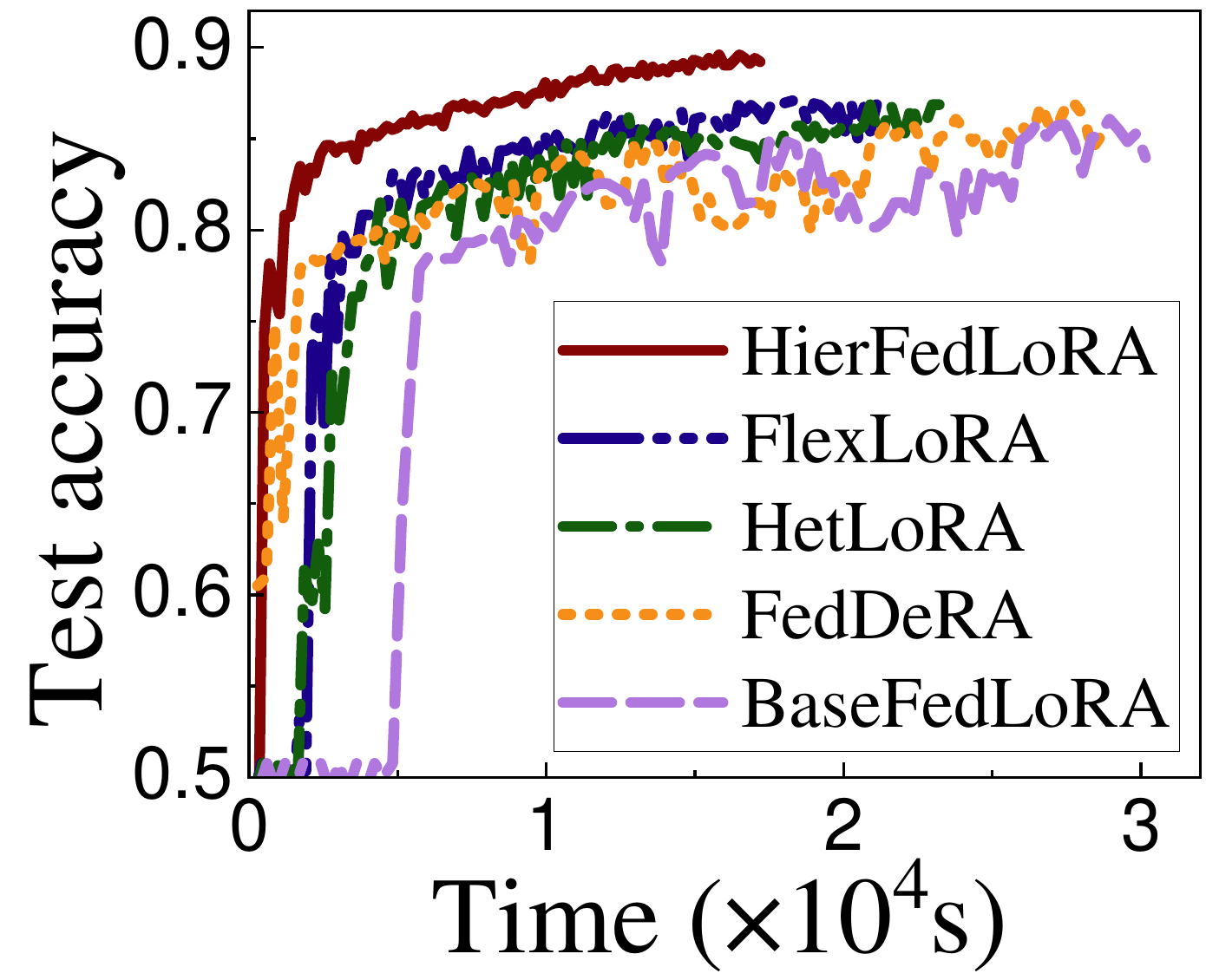}
    		\end{minipage}
		      \label{time-acc-non-iid-qnli}
    	}
	% \\ 
	\subfigure[QQP]{
		\begin{minipage}[b]{0.23\textwidth}
			\includegraphics[width=1\textwidth]{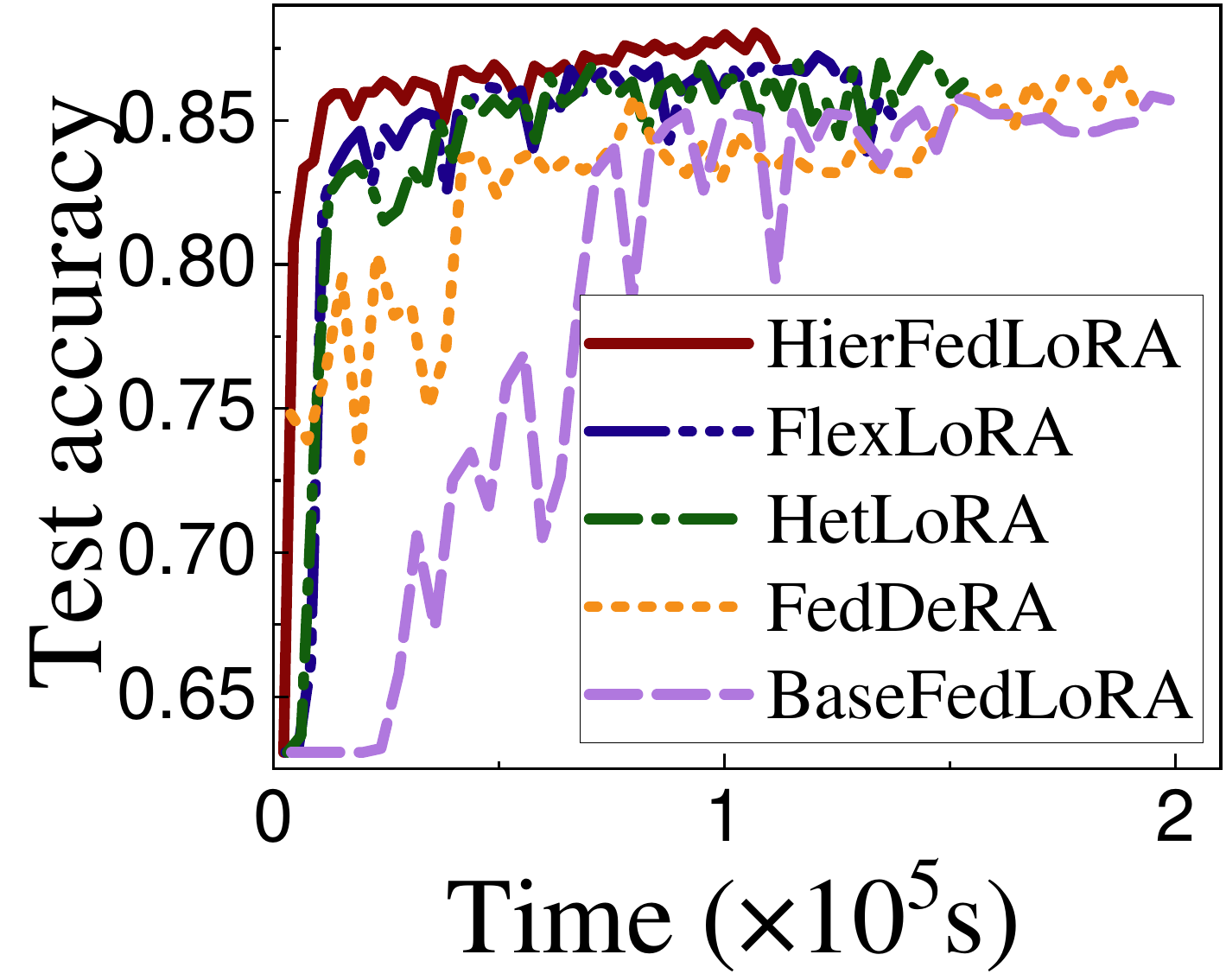}
		\end{minipage}
		\label{time-acc-non-iid-qqp}
	}
        \subfigure[MNLI]{
            \begin{minipage}[b]{0.23\textwidth}
            \includegraphics[width=1\textwidth]{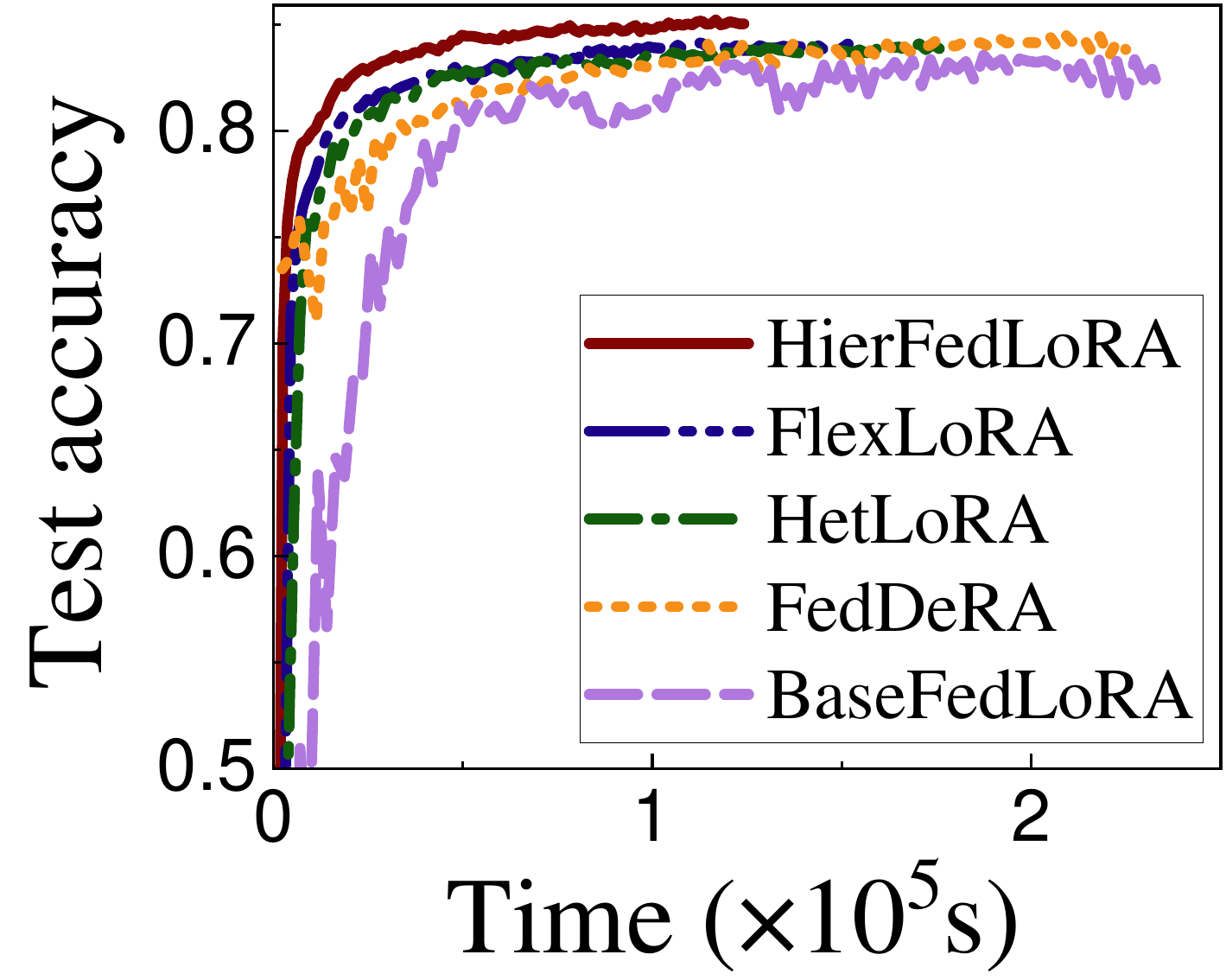}
            \end{minipage}
        \label{time-acc-non-iid-mnli}
        }
        \vspace{-0.4cm}
	\caption{Time-to-accuracy of five approaches on the four non-IID datasets.}
        \vspace{-0.3cm}
        \label{time-acc-non-iid}
\end{figure*}

\textbf{Metrics.}
The following metrics are adopted to evaluate the performance of \oursys and the baselines.

1) \textbf{\textit{Test Accuracy}} reflects the accuracy of the LLMs fine-tuned by different approaches on the test datasets, measured by the proportion of correctly predicted data.
Specifically, we record the test accuracy of the global model (the model after aggregation at the PS) in each round.

2) \textit{\textbf{Time-to-Accuracy}} represents the total wall-clock time required for fine-tuning an LLM to achieve a target accuracy (\ie, fine-tuning time). 
For fair comparisons, we set the target accuracy as the minimum accuracy achieved by the four methods. 
We record the completion time of each round, summing it up to obtain the total fine-tuning time, and also record the average waiting time to reflect the fine-tuning efficiency of different approaches.

3) \textit{\textbf{Communication Traffic}} is recorded by summing up the network traffic for transmitting the tunable parameters between the PS and devices during the process of federated fine-tuning, which is used to measure the communication efficiency of each approach.

\textbf{Experimental Parameters.}
By default, each set of experiments will run 100 rounds.
% We adopt gradient accumulation \cite{} to alleviate the on-device memory constraint.
% For SST-2, the batch size is 16 and the gradient accumulation step is specified as 2.
% The batch size and the gradient accumulation step for QNLI, QQP, and MNLI are identical and are set as 4 and 4, respectively \cite{}.
For SST-2, the batch size is set as 16, and the maximum sequence length of the input text is specified as 256. 
The batch size and max sequence length for QNLI, QQP, and MNLI are identical, which are set as 4 and 512, respectively.
Each device fine-tunes locally for 1 epoch per round using the AdamW optimizer \cite{loshchilov2017decoupled}, and the learning rate is 5e-4, which decays according to a cosine scheduler.
\vspace{-0.2cm}

\subsection{Overall Performance}
Firstly, we conduct sets of experiments on the IID datasets to evaluate the performance of \oursys and the baselines.
The fine-tuning processes of the five approaches are presented in Figs \ref{time-acc-iid}.
According to the results, all the approaches achieve similar test accuracy eventually on the four datasets.
However, \oursys achieves the fastest convergence, followed by FlexLoRA, which is much faster than the other approaches on all four datasets.
For example, as illustrated in Fig. \ref{time-acc-iid-sst2}, \oursys takes 2,833s to achieve 95.5\% test accuracy for RoBERTa on SST-2, while FlexLoRA, HetLoRA, FedDeRA, and BaseFedLoRA consume 3,348s, 3,825s, 4,827s, and 5,213s, respectively.
Similarly, by Fig. \ref{time-acc-iid-qnli}, for RoBERTa on QNLI, when achieving a target accuracy of 91\%, \oursys can separately speed up training by about 1.35$\times$, 1.87$\times$, 2.01$\times$, and 2.13$\times$, compared with FlexLoRA, HetLoRA, FedDeRA, and BaseFedLoRA.
Both FlexLoRA and HetLoRA have adaptive and diverse ranks for heterogeneous devices to speed up the fine-tuning process, while \oursys has optimized device grouping and fine-tuning depth optimization to overcome the challenge of resource constraints.
Specifically, for DeBERTa on QQP, as shown in Fig. \ref{time-acc-iid-qqp}, \oursys reduces the total fine-tuning time by about 33.4\%, 40.3\%, 52.2\%, and 54.1\%, compared to the baselines (\ie, FlexLoRA, HetLoRA, FedDeRA, and BaseFedLoRA).
In addition, by Fig. \ref{time-acc-iid-mnli}, \oursys takes 108,028s to achieve 85.6\% test accuracy for DeBERTa on MNLI, while FlexLoRA, HetLoRA, FedDeRA, and BaseFedLoRA separately consume 137,620s, 159,901s, 188,933s, and 189,777s.

\begin{figure*}
	\centering
    % \hspace{3.5mm}
	\subfigure[SST-2]{
		\begin{minipage}[b]{0.23\textwidth}
			\includegraphics[width=1\textwidth]{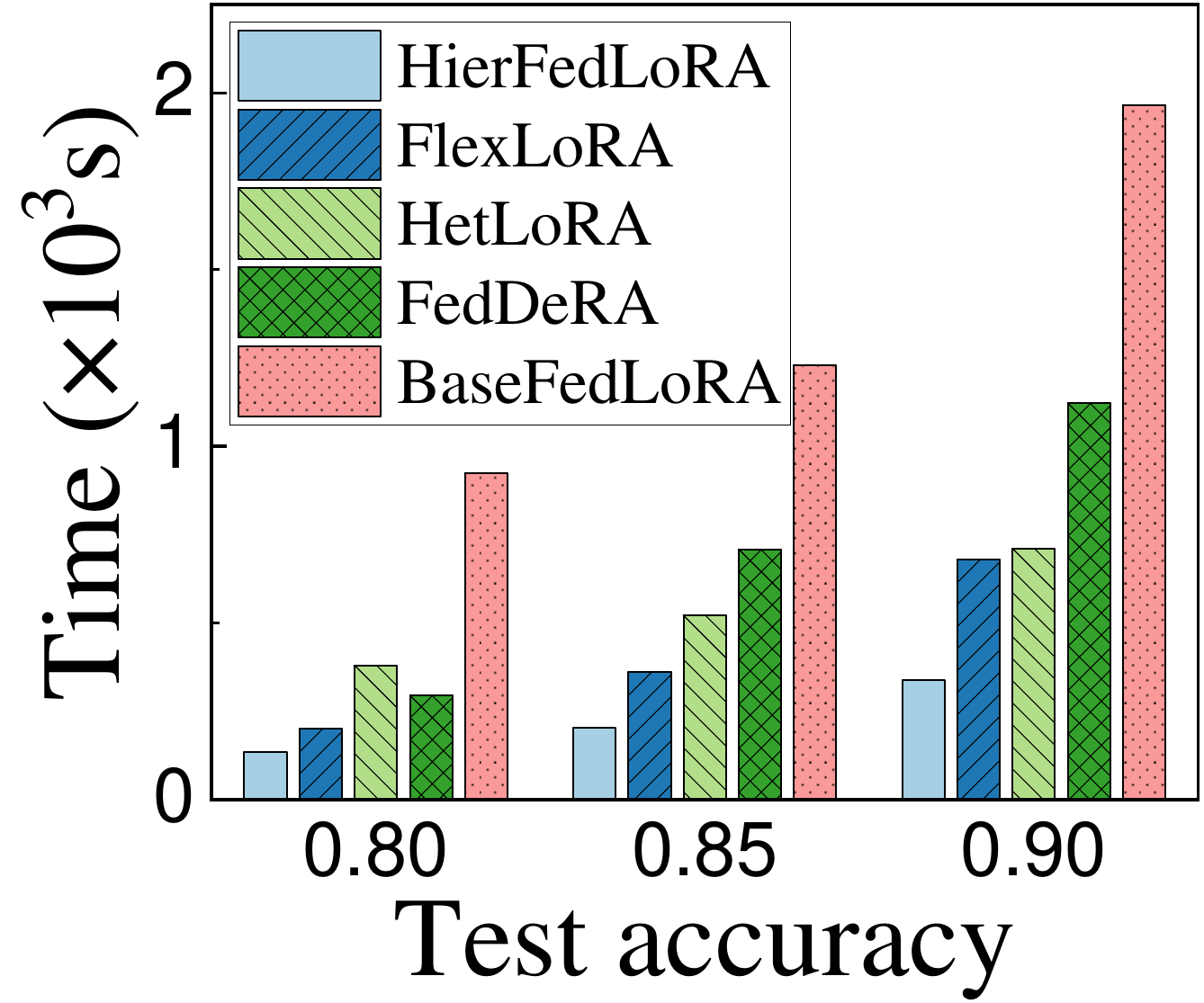}
		\end{minipage}
		\label{completion-time-non-iid-sst2}
	}
    	\subfigure[QNLI]{
    		\begin{minipage}[b]{0.23\textwidth}
   		 	\includegraphics[width=1\textwidth]{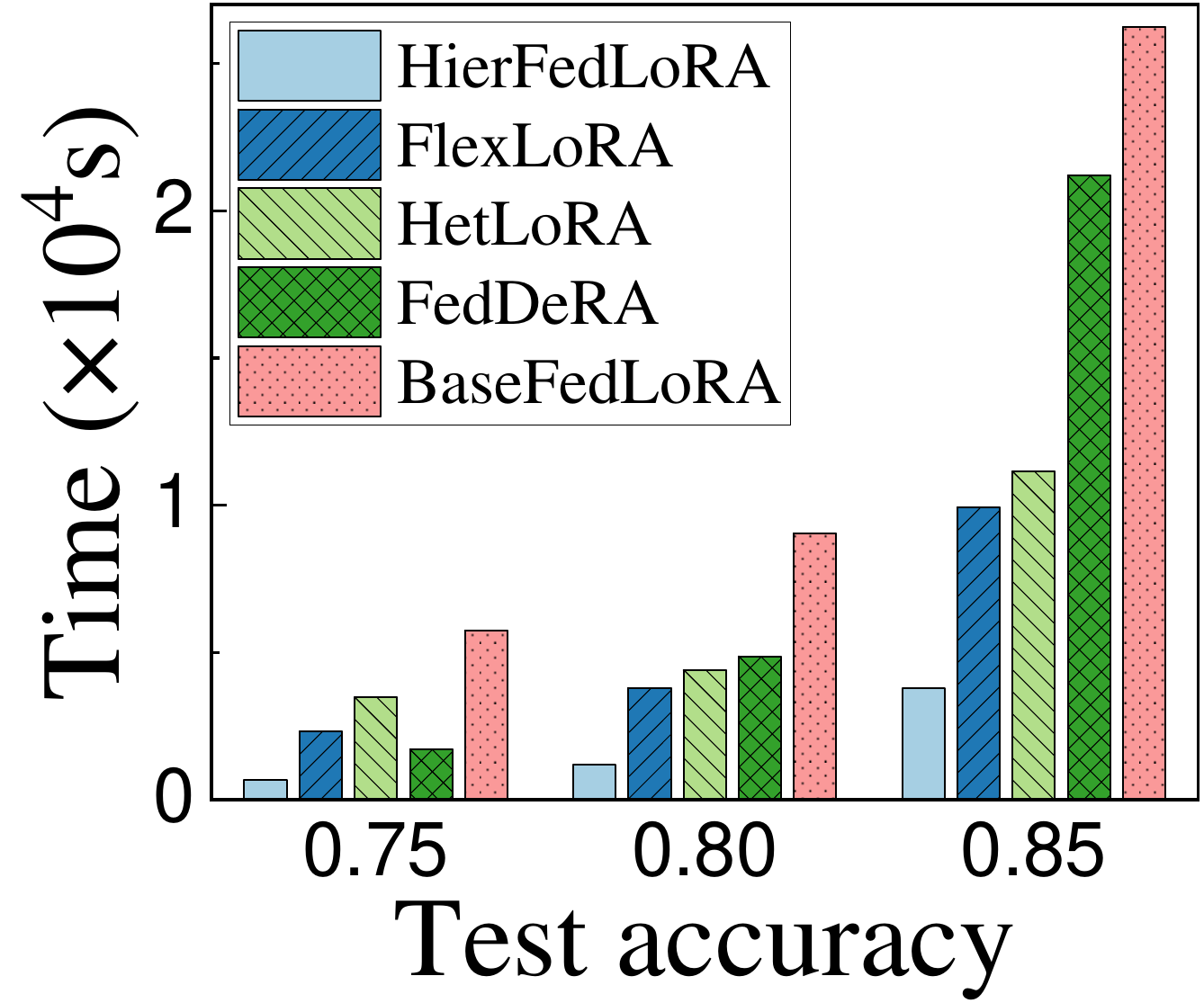}
    		\end{minipage}
		\label{completion-time-non-iid-qnli}
    	}
	% \\ 
	\subfigure[QQP]{
		\begin{minipage}[b]{0.23\textwidth}
			\includegraphics[width=1\textwidth]{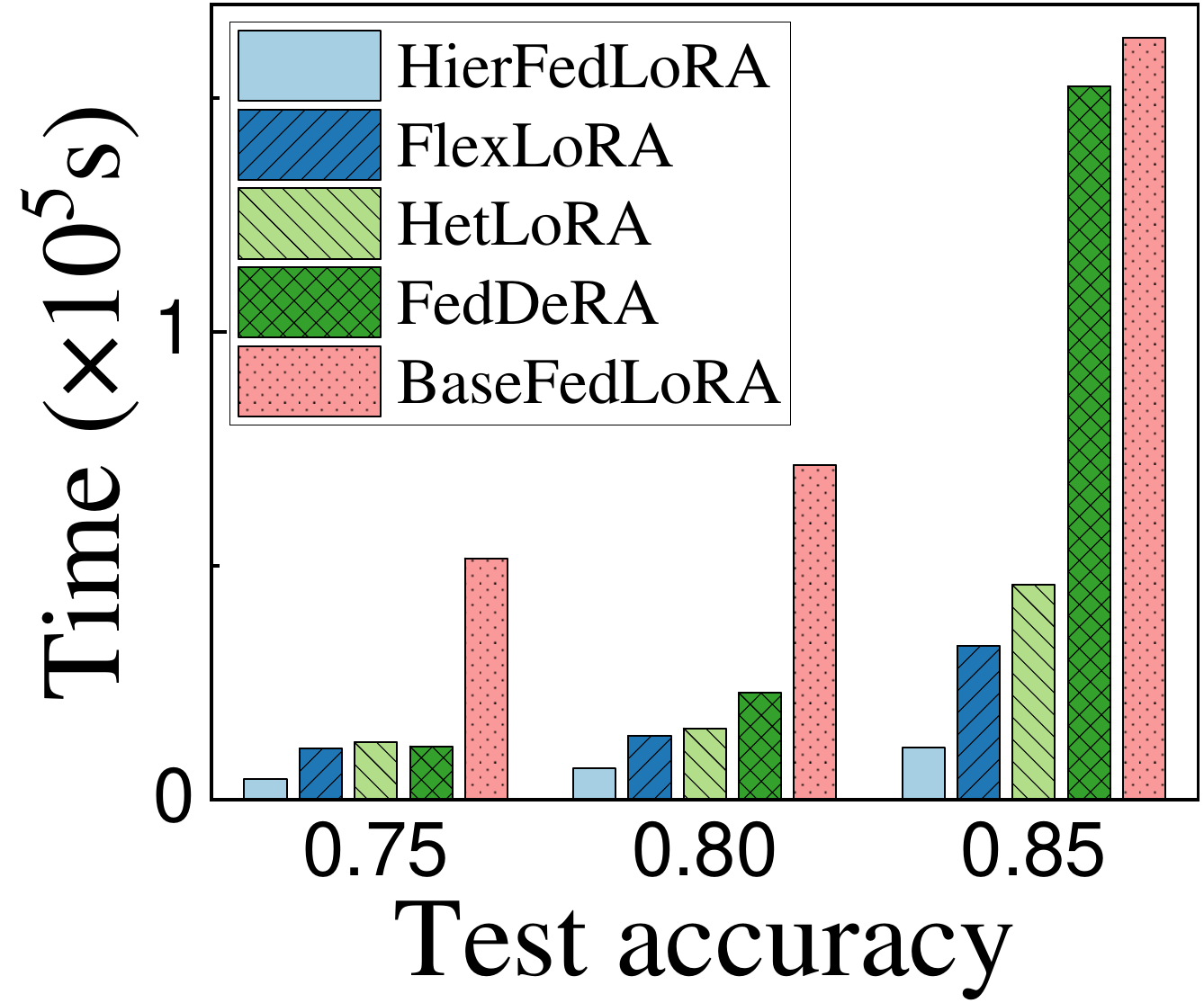}
		\end{minipage}
		\label{completion-time-non-iid-qqp}
	}
    	\subfigure[MNLI]{
    		\begin{minipage}[b]{0.23\textwidth}
		 	\includegraphics[width=1\textwidth]{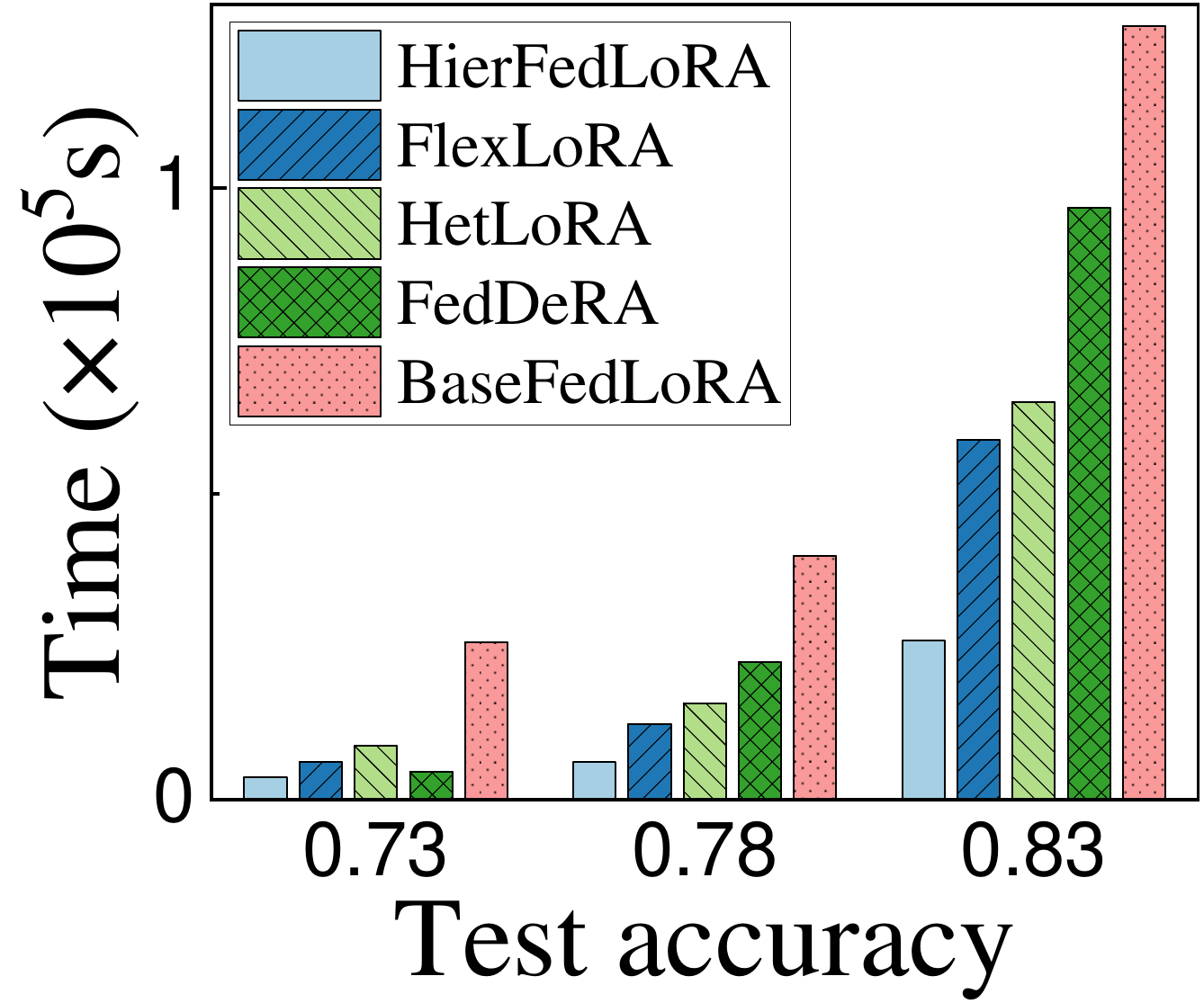}
    		\end{minipage}
		      \label{completion-time-non-iid-mnli}
    	}
        \vspace{-0.4cm}
	\caption{Completion time of five approaches when achieving different target accuracies.}
        \vspace{-0.3cm}
        \label{completion-time-non-iid}
\end{figure*}

\begin{figure*}
	\centering
    % \hspace{3.5mm}
	\subfigure[SST-2]{
		\begin{minipage}[b]{0.23\textwidth}
			\includegraphics[width=1\textwidth]{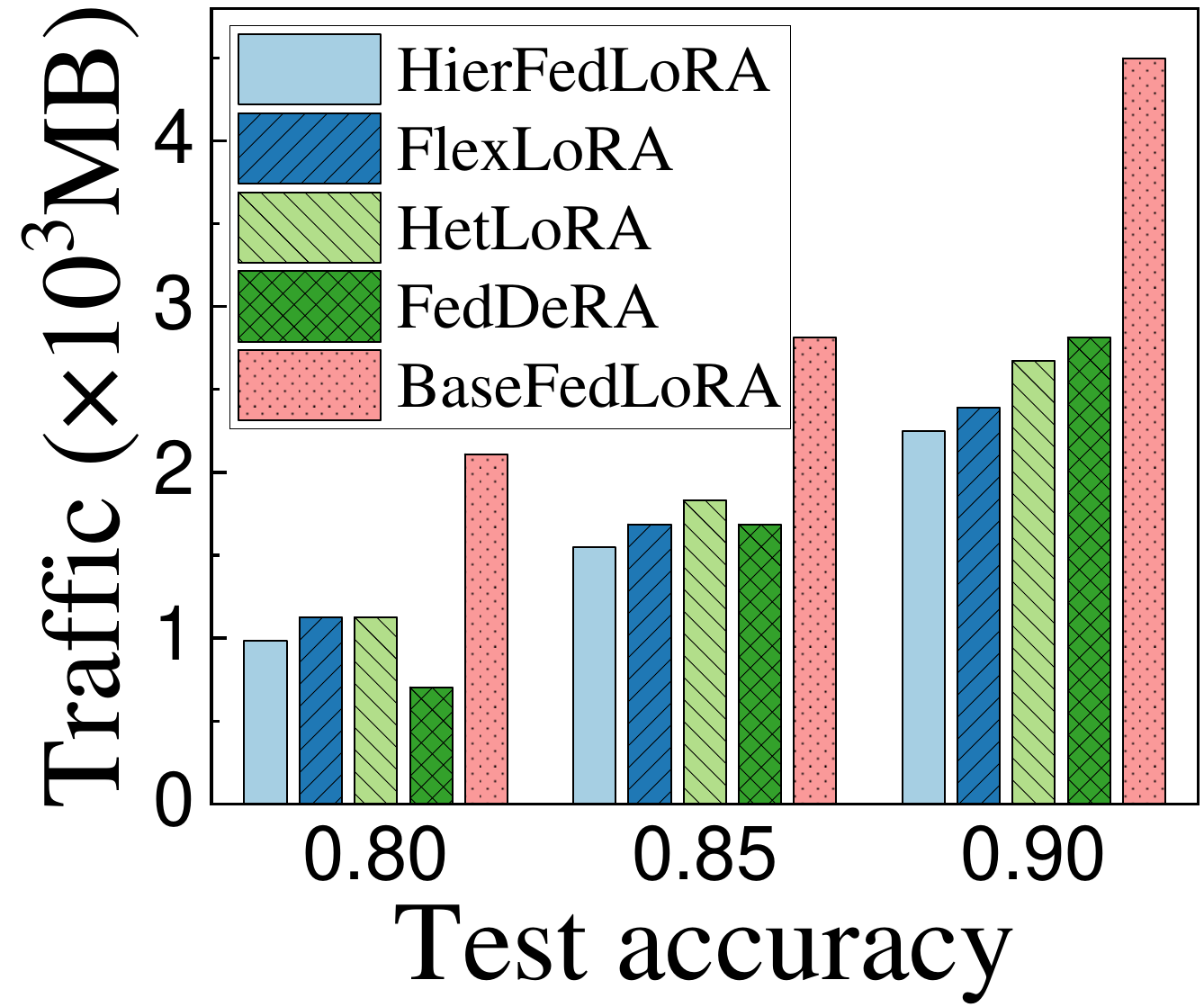}
		\end{minipage}
		\label{network-traffic-diff-non-iid-sst2}
	}
        \subfigure[QNLI]{
            \begin{minipage}[b]{0.23\textwidth}
            \includegraphics[width=1\textwidth]{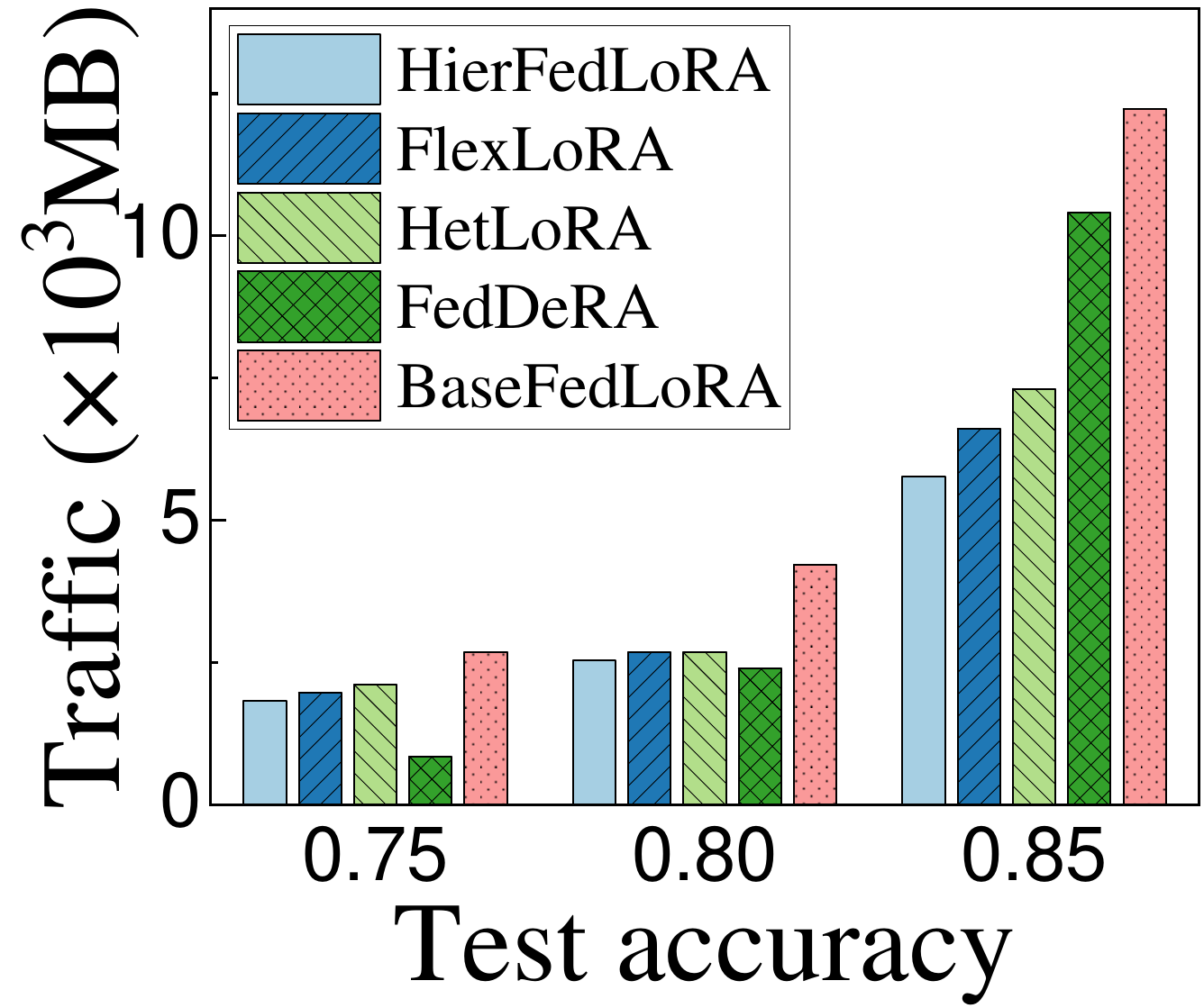}
            \end{minipage}
            \label{network-traffic-diff-non-iid-qnli}
        }
    	% \\ 
	\subfigure[QQP]{
		\begin{minipage}[b]{0.23\textwidth}
			\includegraphics[width=1\textwidth]{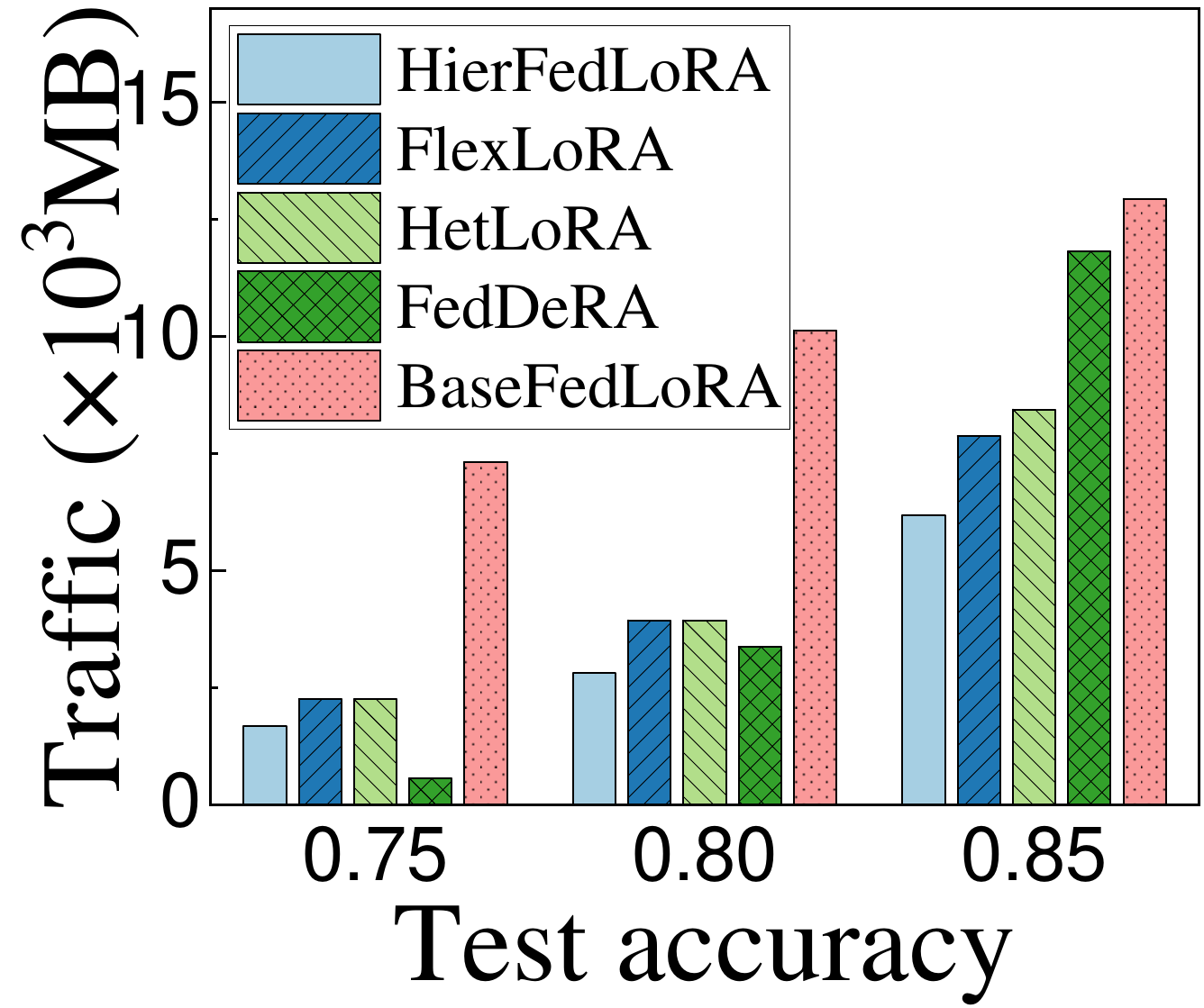}
		\end{minipage}
		\label{network-traffic-diff-non-iid-qqp}
	}
        \subfigure[MNLI]{
            \begin{minipage}[b]{0.23\textwidth}
            \includegraphics[width=1\textwidth]{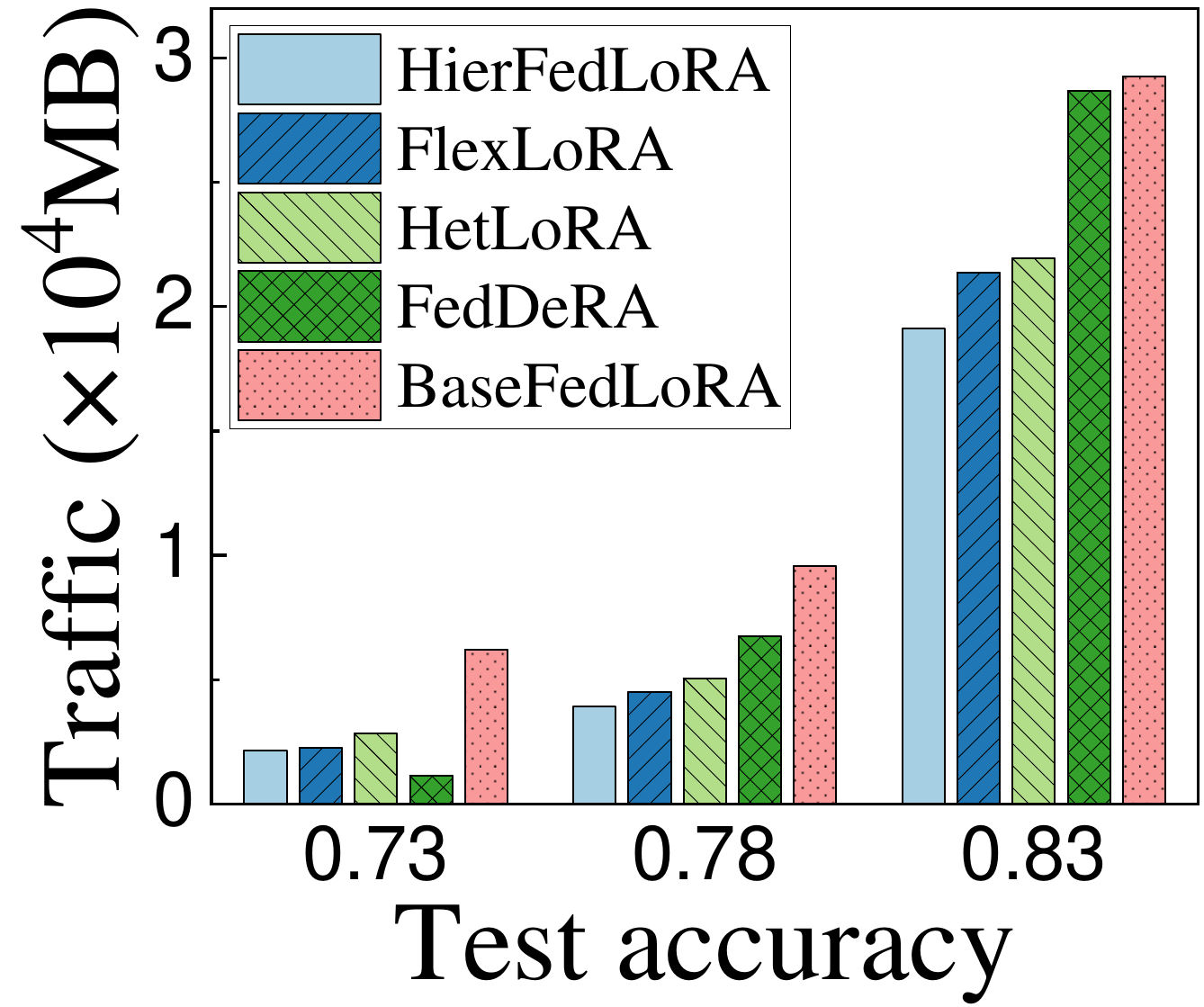}
            \end{minipage}
            \label{network-traffic-diff-non-iid-mnli}
        }
        \vspace{-0.4cm}
	\caption{Network traffic of five approaches when achieving different target accuracies.}
        \vspace{-0.3cm}
        \label{network-traffic-diff-non-iid}
\end{figure*}

Secondly, we also conduct a set of experiments of these approaches on the four datasets with non-IID level $p=10$, and the results are presented in Figs \ref{time-acc-non-iid} and \ref{completion-time-non-iid}.
We find that all the approaches maintain a similar convergence rate as that in the IID setting but suffer from varying degrees of accuracy degradation.
However, \oursys with adaptive device grouping and depth optimization achieves the highest accuracy among these approaches.
For instance, as shown in Fig. \ref{time-acc-non-iid-sst2}, \oursys achieves 94.8\% accuracy in 2,933s for RoBERTa on SST-2, while FlexLoRA, HetLoRA, FedDeRA, and BaseFedLoRA takes 3,984s, 4,720s, 5885s, and 6,130s to reach the accuracy of 93.7\%, 93.6\%, 93.6\%, and 92.1\%, respectively.
Similarly, as illustrated in Fig. \ref{time-acc-non-iid-qnli}, for RoBERTa on QNLI with the same fine-tuning time of 15,000s, \oursys improves the test accuracy by about 3.1\%, 4.2\%, 4.6\%, and 5.2\%, compared to FlexLoRA, HetLoRA, FedDeRA, and BaseFedLoRA, respectively.
FedDeRA with intuitive LoRA initialization improves the fine-tuning process to some extent, while FlexLoRA and HetLoRA with adaptive rank for the devices speed up the convergence and slightly improve test accuracy compared to the BaseFedLoRA.
Specifically, according to the results in Fig. \ref{time-acc-non-iid-qqp}, \oursys separately improves the final accuracy by about 0.8\%, 1.1\%, 1.7\%, and 3.1\% for DeBERTa on QQP, compared to FlexLoRA, HetLoRA, FedDeRA, and BaseFedLoRA, respectively.
Besides, by Figs \ref{time-acc-non-iid-mnli} and \ref{completion-time-non-iid-mnli}, when achieving 83\% test accuracy, \oursys takes 26,084s for DeBERTa on MNLI, while FlexLoRA, HetLoRA, FedDeRA, and BaseFedLoRA takes 58,769s, 65,027s, 96,721s, and 119,485s, respectively.
These results show that \oursys is effective in addressing data heterogeneity.

Thirdly, to further illustrate the advantage of \oursys in saving communication resources, we illustrate the completion time and network traffic of these approaches when achieving different target accuracies in Fig. \ref{network-traffic-diff-non-iid}, respectively. 
According to the results, the network traffic consumption of all approaches increases with the target accuracy for all four datasets.
\oursys always consumes the least network traffic among all approaches.
This is because \oursys leverages the advantage of hierarchical aggregation and adaptive fine-tuning depth, speeding up the fine-tuning process and reducing the network traffic.
Specifically, as shown in Fig. \ref{network-traffic-diff-non-iid-sst2}, when achieving 90\% accuracy, \oursys, FlexLoRA, and HetLoRA consume 2,250MB, 2,390MB, and 2,671MB, respectively, while FedDeRA and BaseFedLoRA consume 2,813MB and 4,502MB for RoBERTa on SST-2.
By Fig. \ref{network-traffic-diff-non-iid-qnli}, for fine-tuning RoBERTa on QNLI to achieve 85\% accuracy, \oursys saves network traffic consumption by about 12.8\%, 21.2\%, 44.6\%, and 52.9\%, compared to FlexLoRA, FedDeRA, HetLoRA, and BaseFedLoRA, respectively. 
Besides, as illustrated in Fig. \ref{network-traffic-diff-non-iid-qqp}, \oursys reduces the network traffic consumption by about 1,687MB, 2,253MB, 5,625MB, and 6,758MB when achieving 85\% accuracy for DeBERTa on QQP, compared to the baselines (\ie, FlexLoRA, HetLoRA, FedDeRA, and BaseFedLoRA). 
Moreover, by Fig. \ref{network-traffic-diff-non-iid-mnli}, for fine-tuning DeBERTa on MNLI to achieve 83\% accuracy, \oursys reduces the network traffic consumption by about 11.2\%, 12.8\%, 33.5\%, and 34.6\%, compared to FlexLoRA, HetLoRA, FedDeRA, and BaseFedLoRA, respectively.

\begin{figure*}
    \centering
    \hspace{-3.5mm}
	\subfigure[SST-2]{
		\begin{minipage}[b]{0.23\textwidth}
			\includegraphics[width=1\textwidth]{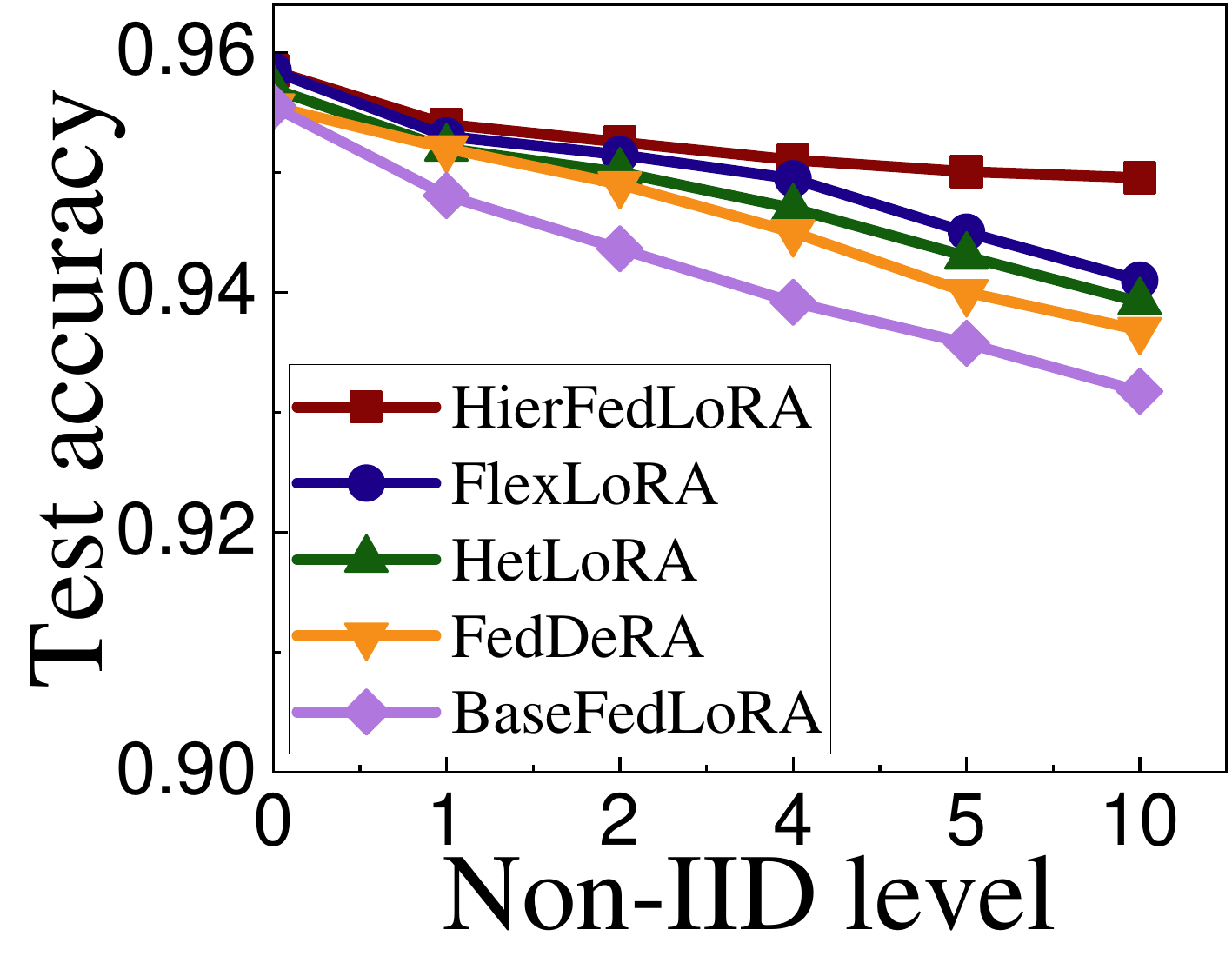}
		\end{minipage}
		\label{final-acc-diff-non-iid-sst2}
	}
        \subfigure[QNLI]{
            \begin{minipage}[b]{0.23\textwidth}
            \includegraphics[width=1\textwidth]{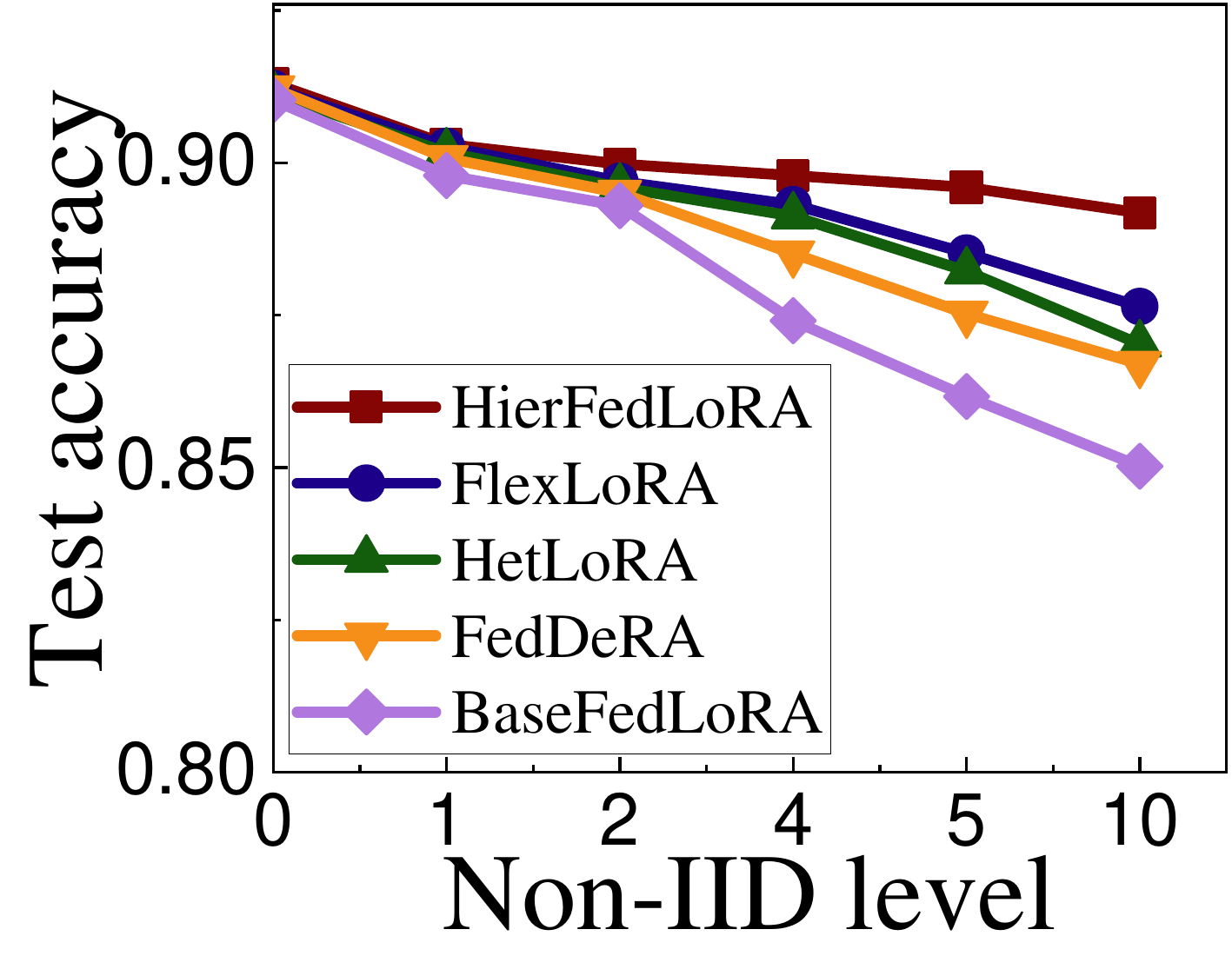}
            \end{minipage}
            \label{final-acc-diff-non-iid-qnli}
        }
	% \\ 
	\subfigure[QQP]{
		\begin{minipage}[b]{0.23\textwidth}
			\includegraphics[width=1\textwidth]{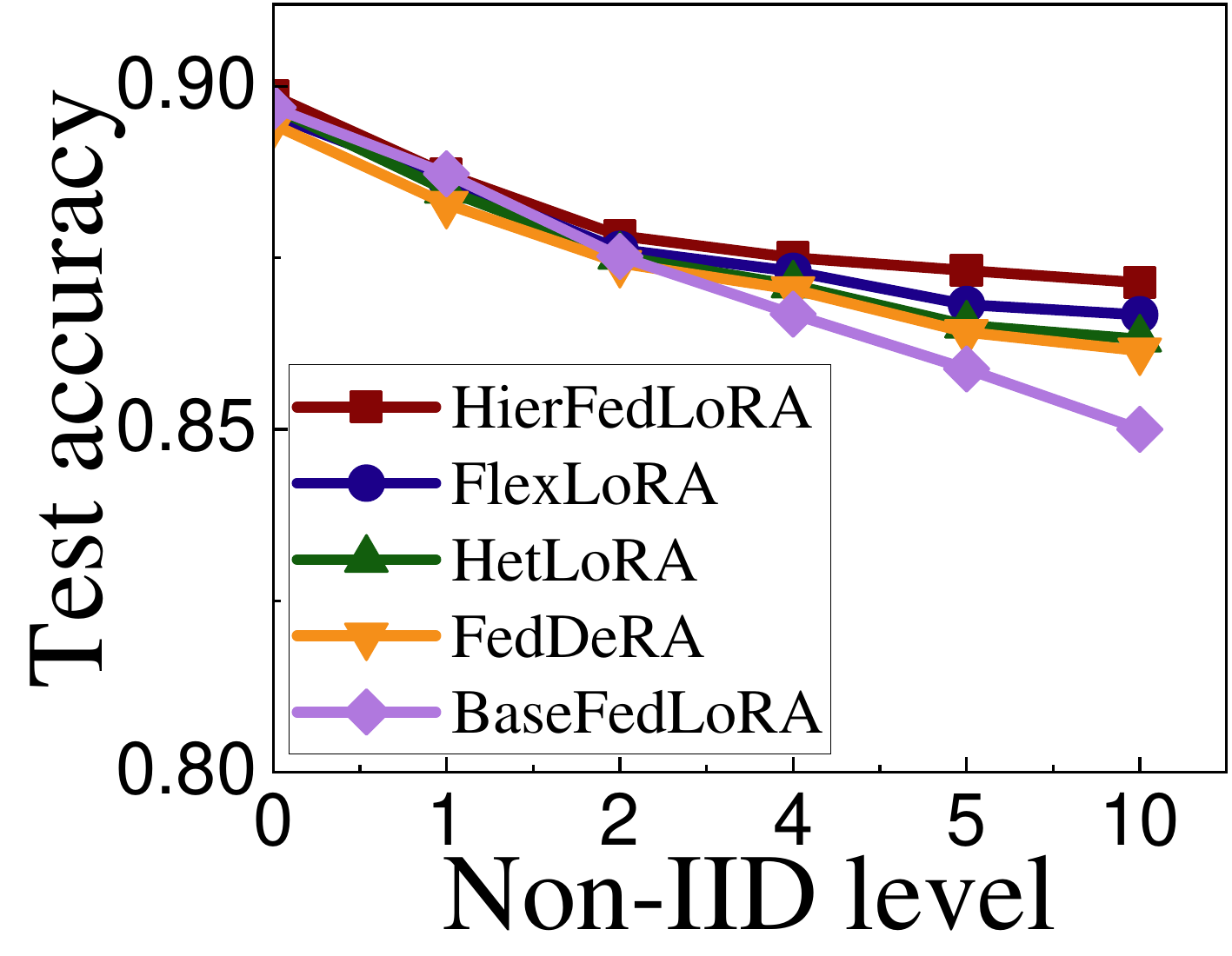}
		\end{minipage}
		\label{final-acc-diff-non-iid-qqp}
	}
        \subfigure[MNLI]{
            \begin{minipage}[b]{0.23\textwidth}
            \includegraphics[width=1\textwidth]{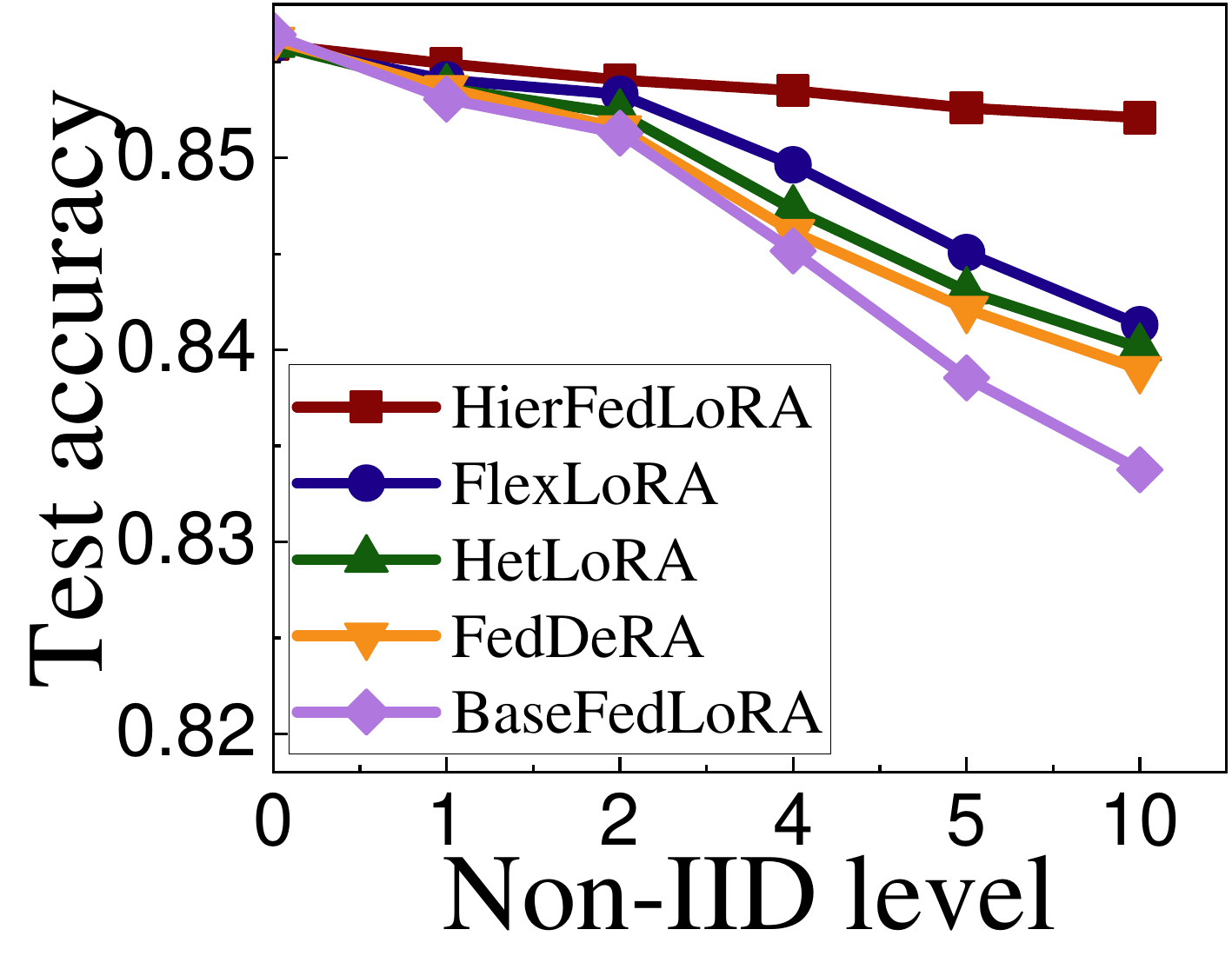}
            \end{minipage}
            \label{final-acc-diff-non-iid-mnli}
        }
        \vspace{-0.6cm}
	\caption{Final accuracy of the five approaches on the four datasets with different non-IID levels.}
        \vspace{-0.5cm}
        \label{final-acc-diff-non-iid}
\end{figure*}

\subsection{Effect of Non-IID Level}
To demonstrate the effectiveness of \oursys in handling non-IID data, we present the test accuracy of different approaches at varying non-IID levels in Fig. \ref{final-acc-diff-non-iid}, in which the model accuracy of the five approaches on all the datasets decreases as the non-IID level increases. 
However, \oursys consistently outperforms the other approaches on all datasets. 
FlexLoRA and HetLoRA, without considering the challenges of data heterogeneity, exhibit the lowest model accuracy on non-IID datasets. 
Specifically, as illustrated in Fig. \ref{final-acc-diff-non-iid-sst2}, \oursys can achieve improvement of test accuracy by about 0.9\%, 1.0\%, 1.3\%, and 1.8\% on SST-2 with the non-IID level of $p=10$, compared to the baselines (\ie, FlexLoRA, HetLoRA, FedDeRA, and BaseFedLoRA). 
Notably, by Fig. \ref{final-acc-diff-non-iid-qnli}, with the non-IID level of $p=10$ on QNLI, \oursys achieves improvement of final accuracy by about 1.5\%, 2.2\%, 2.5\%, and 4.2\%, compared to the baselines (\ie, FlexLoRA, HetLoRA, FedDeRA, BaseFedLoRA). 
Besides, as shown in Fig. \ref{final-acc-diff-non-iid-qqp}, while transitioning from IID to non-IID level of $p=10$ on QQP, \oursys, FlexLoRA, HetLoRA, and FedDeRA suffer from only 2.7\%, 3.2\%, 3.5\%, and 3.7\% loss in accuracy, while the accuracy loss for BaseFedLoRA is 4.9\%. 
Moreover, by Fig. \ref{final-acc-diff-non-iid-mnli}, with the non-IID level of $p=10$ on MNLI, \oursys, FlexLoRA, HetLoRA, and FedDeRA achieve 85.2\%, 84.1\%, 84.0\%, and 83.9\% accuracy, while BaseFedLoRA only achieves 83.4\%. 
These results further demonstrate the advantage of \oursys in addressing data heterogeneity by aggregation frequency and depth optimization.

\begin{figure}[t]
    \centering
    \hspace{-1mm}
    \subfigure[QNLI (IID)]{
        \begin{minipage}[t]{0.5\linewidth}
        \centering
        \includegraphics[width=1.7in]{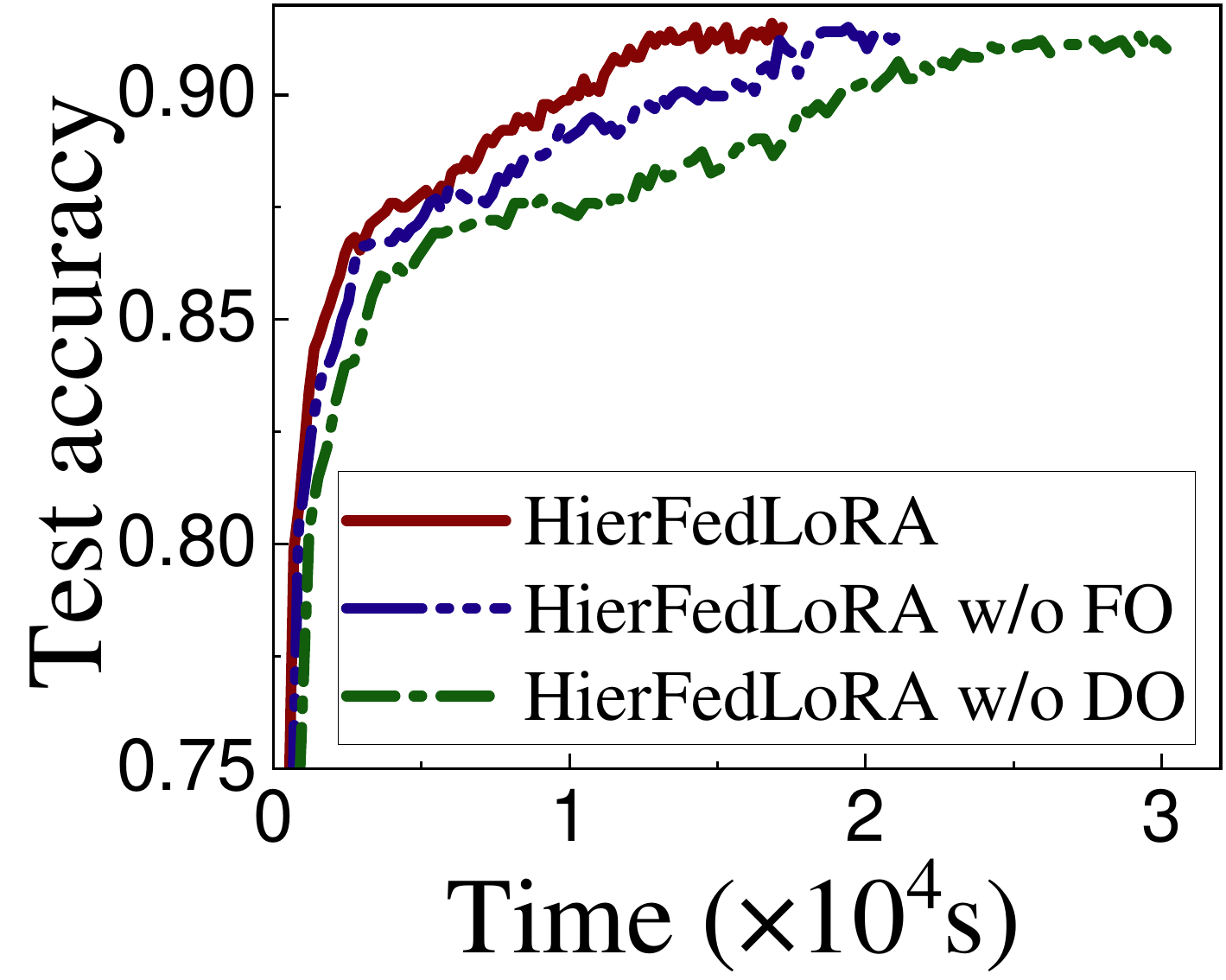}
        \end{minipage}%
        \label{ablation-qnli-iid}
    }%
    \subfigure[QNLI (non-IID)]{
        \begin{minipage}[t]{0.5\linewidth}
        \centering
        \includegraphics[width=1.7in]{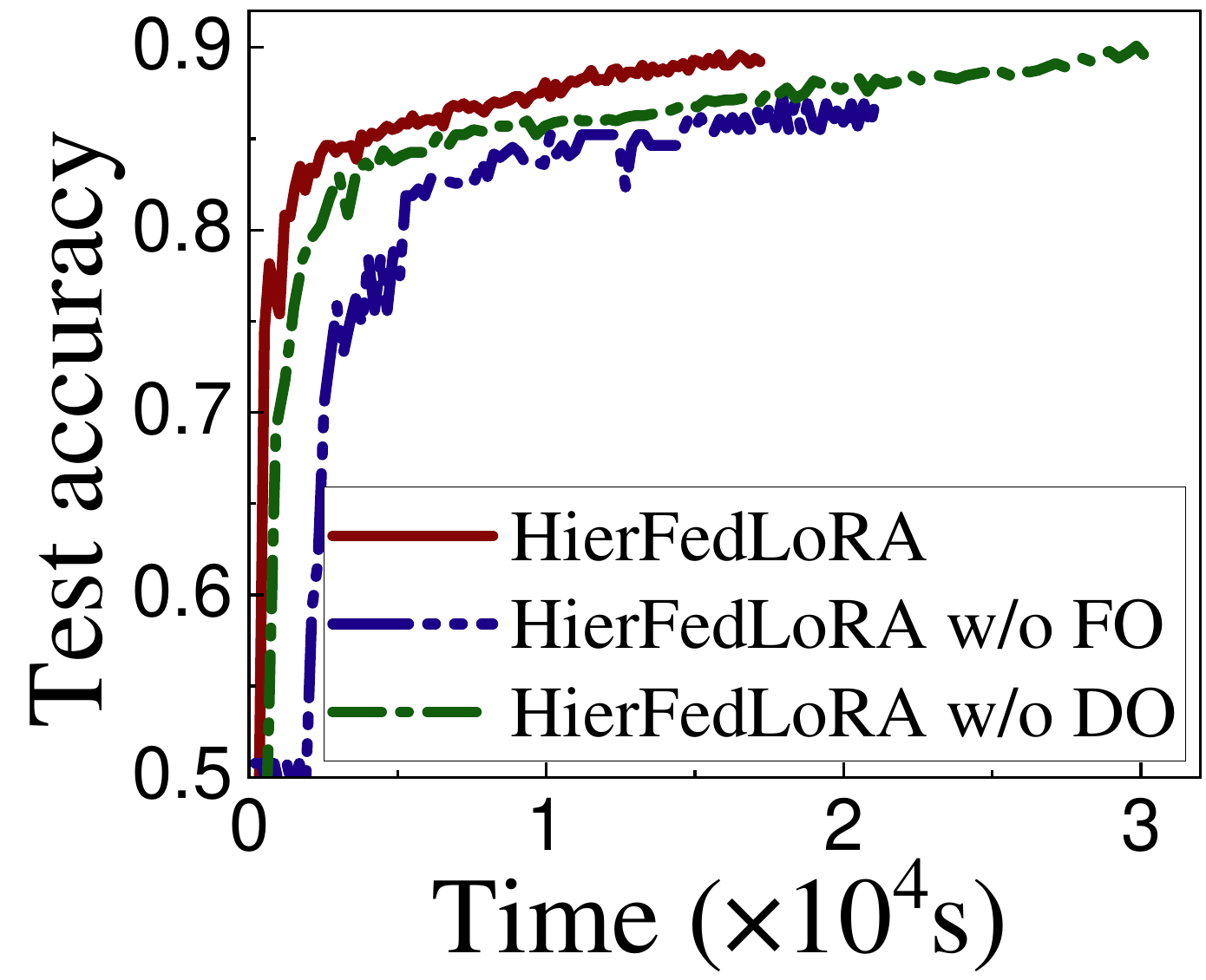}
        \end{minipage}%
        \label{ablation-qnli-non-iid}
    }%
    \centering
    \vspace{-0.5cm}
    \caption{Effect of key strategies.}
    \label{ablation-qnli}
    \vspace{-0.4cm}
\end{figure}

\subsection{Effect of Key Strategies}
There are two key strategies of \oursys, \ie, aggregation frequency optimization and fine-tuning depth adaptation, which are developed to enhance the performance of vanilla FedLoRA.
Herein, we conduct several sets of experiments for fine-tuning RoBERTa on QNLI with IID distribution ($p=0$) and non-IID distribution ($p=10$) to evaluate the effectiveness of the two critical strategies.
We adopt the \oursys without frequency optimization (\oursys w/o FO) and typical FedLoRA without depth optimization (\oursys w/o DO) as the baselines. 
Concretely, in \oursys w/o FO, the PS assigns the identical and fixed aggregation frequency (\ie, $\rho = 1$) for each group, while in \oursys w/o DO, the PS sets each group with the same depth (\ie, $d = 12$). 
By Fig. \ref{ablation-qnli}, \oursys w/o FO converges much faster than \oursys w/o DO on the IID dataset, and \oursys w/o FO suffers from accuracy degradation than \oursys w/o DO on the non-IID dataset. 
Specifically, the \oursys w/o FO degrades the final test accuracy by about 2.2\% on the non-IID dataset compared to \oursys w/o DO. 
The results illustrate that our aggregation frequency optimization is essential for addressing the data heterogeneity. 
Besides, powered by the fine-tuning depth optimization among device groups, \oursys speeds up fine-tuning by about 1.75$\times$ compared to \oursys w/o DO on the non-IID settings. 
The results demonstrate the positive roles of aggregation frequency and depth optimization in \oursys.

\subsection{Effect of System Scale}
\begin{figure}[t]
    \centering
    \hspace{-10mm}
    \subfigure[Completion Time]{
        \begin{minipage}[t]{0.5\linewidth}
        \centering
        \includegraphics[width=1.65in]{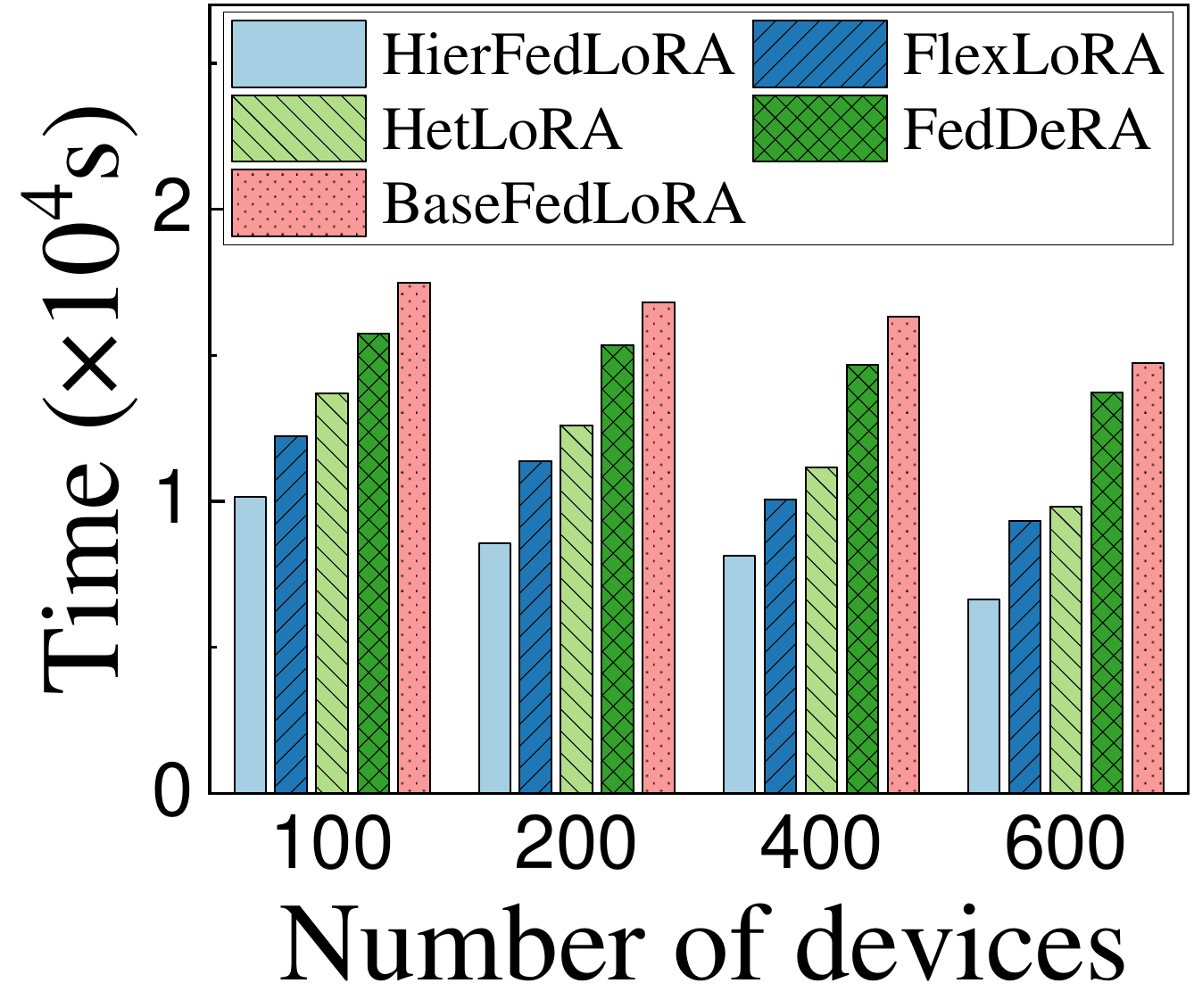}
        \end{minipage}%
        \label{sys-scale-completion}
    }%
    \subfigure[Fine-tuning Process]{
        \begin{minipage}[t]{0.5\linewidth}
        \centering
        \includegraphics[width=1.7in]{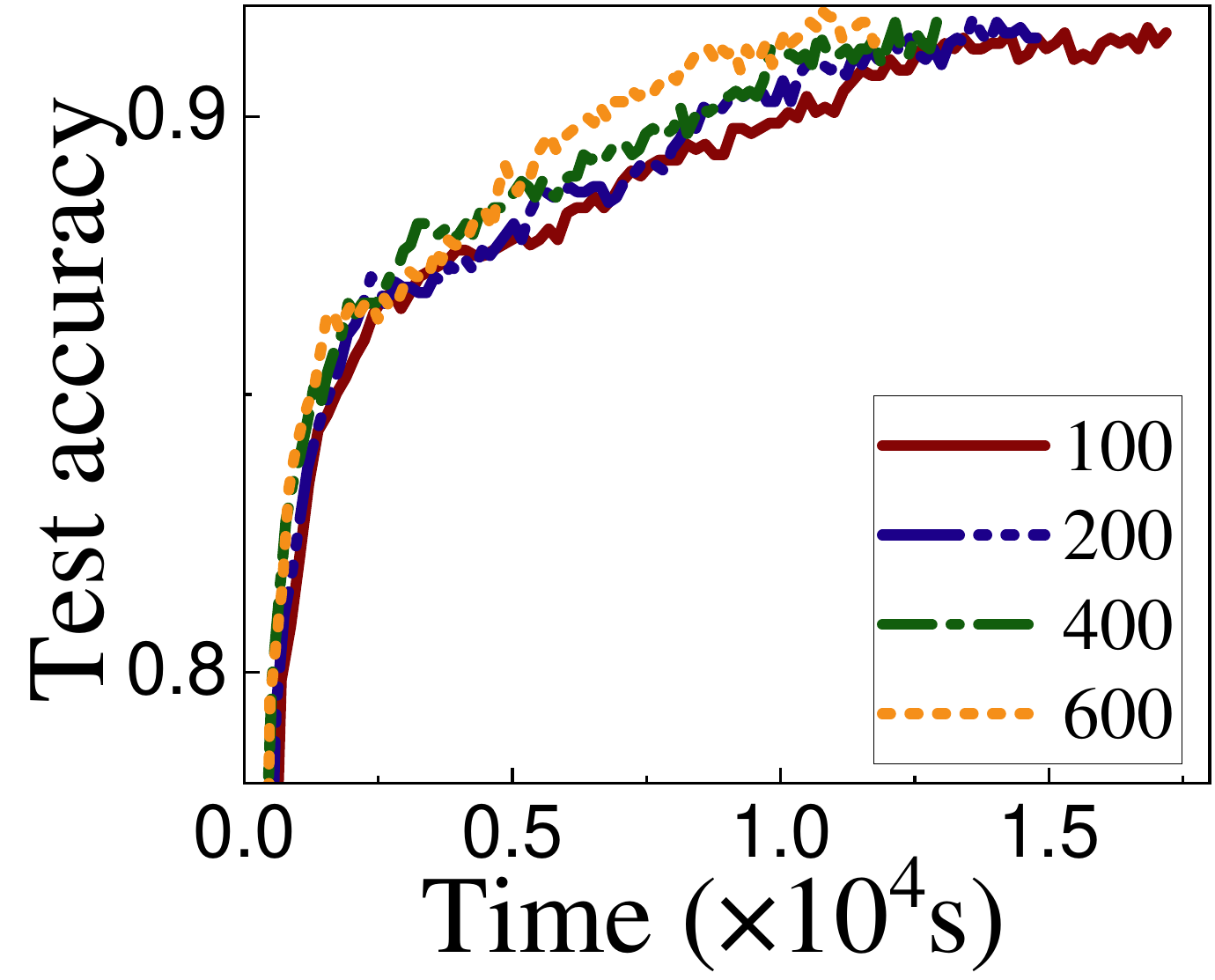}
        \end{minipage}%
        \label{sys-scale-time2acc}
    }%
    \centering
    \vspace{-0.5cm}
    \caption{Effect of system scale.}
    \label{sys-scale}
    \vspace{-0.4cm}
\end{figure}

To demonstrate the robustness of \oursys, we evaluate the performance of \oursys and baselines with different scales of participating devices.
We conduct several sets of experiments for fine-tuning RoBERTa on QNLI with four scales (\eg, 100, 200, 400, 600) through extensive simulation experiments, which are conducted on an AMAX deep learning workstation equipped with an Intel(R) Xeon(R) Platinum 8360Y CPU @ 2.4GHz, 4 NVIDIA A100 GPUs (80GB memory each) and 512 GB RAM.
The results of completion time to achieve 90\% accuracy are presented in Fig. \ref{sys-scale-completion}, and the fine-tuning processes of different scale of \oursys are presented in Fig. \ref{sys-scale-time2acc}.
As the number of participating devices increases, all approaches achieve faster convergence.
This is because the number of data samples on a device is limited, and more devices contribute more data for fine-tuning in each round, thus speeding up the fine-tuning process.
For example, \oursys with 600 devices reduces the total fine-tuning time by about 31\%, 24\%, and 14\% compared to \oursys with 100, 200, and 400 devices, respectively.
Besides, \oursys also achieves a speedup of 1.29$\times$-2.22$\times$ when reaching 90\% accuracy, compared to the baselines (\ie, FlexLoRA, HetLoRA, FedDeRA, BaseFedLoRA) regarding the different scales of participating devices.
The results further illustrate the robustness and advantage of \oursys.

%-------------------------------------------------------------------------------

%-------------------------------------------------------------------------------
% \vspace{-0.1cm}
% \section{Related Works}\label{sec:relatedwork}
% \input{contents/related_word}

%-------------------------------------------------------------------------------

%-------------------------------------------------------------------------------
% \vspace{-0.1cm}
\section{Conclusion}\label{sec:concluesion}
In this paper, we propose a hierarchical FedLoRA framework, called \oursys, to address data heterogeneity and resource constraints through an effective combination of aggregation frequency and depth adaptation.
% we review the characteristic properties of LoRA and propose an efficient LoRA-based FedLLM framework, called \oursys, to address resource constraints and data heterogeneity.
We develop an efficient algorithm to carefully determine the frequency and depth, aiming to balance the trade-off between fine-tuning efficiency and model performance. 
% We conduct the study of \oursys on a real platform of 80 wireless devices. 
% The experimental results show that \oursys improves the final model accuracy by 1.6\% to 4.2\%, speeding up the fine-tuning process by at least 2.1$\times$, compared to the baselines.
Extensive experiments are conducted on a physical platform with 80 commercial devices. 
The results show that \oursys improves the final model accuracy by 1.6\% to 4.2\%, speeding up the fine-tuning process by at least 2.1$\times$, compared to the strong baselines.

% \section{Notations(temporary)}\label{sec:notation}
% \input{content/notation}
 

% \begin{acks}
% \end{acks}

\balance
\bibliographystyle{unsrt}
\bibliography{contents/refs}

\appendix

\end{document}